\documentclass[10pt,twocolumn,letterpaper]{article}
\usepackage[pagenumbers]{iccv} 

\usepackage[utf8]{inputenc} 
\usepackage[T1]{fontenc}    

\usepackage{url}            
\usepackage{booktabs}       
\usepackage{amsfonts}       
\usepackage{nicefrac}       
\usepackage{microtype}      
\usepackage{lipsum}
\usepackage{fancyhdr}       
\usepackage{graphicx}       
\usepackage{float}
\usepackage{xcolor}
\graphicspath{{media/}}     
\usepackage{amsmath}
\usepackage{tikz}
\usepackage{graphicx}  
\usetikzlibrary{arrows.meta,positioning,calc,shapes}
\usepackage{algorithm}
\usepackage{algorithmic}
\usepackage{placeins}
\usepackage{multirow} 
\usepackage[labelfont=bf,hypcap=true]{caption}
\usepackage{newfloat}
\usepackage{pifont}  
\usepackage{arydshln}  
\usepackage{url}
\usepackage{xurl}
\usepackage{lineno}

\definecolor{iccvblue}{rgb}{0.21,0.49,0.74}
\usepackage[pagebackref,breaklinks,colorlinks,allcolors=iccvblue]{hyperref}
\usepackage[capitalize]{cleveref}

\newcommand{\bx}{\mathbf{x}}
\newcommand{\bxt}{\mathbf{\widehat x}}

\newcommand{\bm}{\mathbf{m}}

\newcommand{\bL}{\mathbf{L}}

\newcommand{\True}{\ding{51}}  

\usepackage[most]{tcolorbox}

\newtcolorbox[auto counter]{floatcolorbox}[3][]{
  label=#1,
  title={\small {Box \thetcbcounter: #2}},
  enhanced,
  breakable,
  coltitle=black,
  fontupper=\small\sffamily,
  rounded corners,                 
  boxrule=0.5pt,                     
  drop shadow southeast,           
  colback=#3!3!white,           
  colframe=#3!80!black,         
  colbacktitle=#3!30!white,
  attach boxed title to top left={xshift=2mm, yshift=-2mm},
  boxed title style={colback=#3!20!white,rounded corners,boxrule=0.9pt}
}

\title{\vspace{-3.0em}Flux Already Knows – Activating Subject-Driven \\ Image Generation without Training\vspace{-0.8em}}

\author{Hao Kang\footnote{These authors contributed equally to this work.}, Stathi Fotiadis\footnotemark[1], Liming Jiang, Qing Yan, \\ Yumin Jia, Zichuan Liu, Min Jin Chong, Xin Lu \\
\\ByteDance Intelligent Creation
}

\begin{document}

\twocolumn[{
\maketitle
\vspace{-3em}
\begin{center}
    \makebox[\textwidth][c]{\resizebox{1.11\textwidth}{!}{\includegraphics[trim=0 135 0 0, clip]{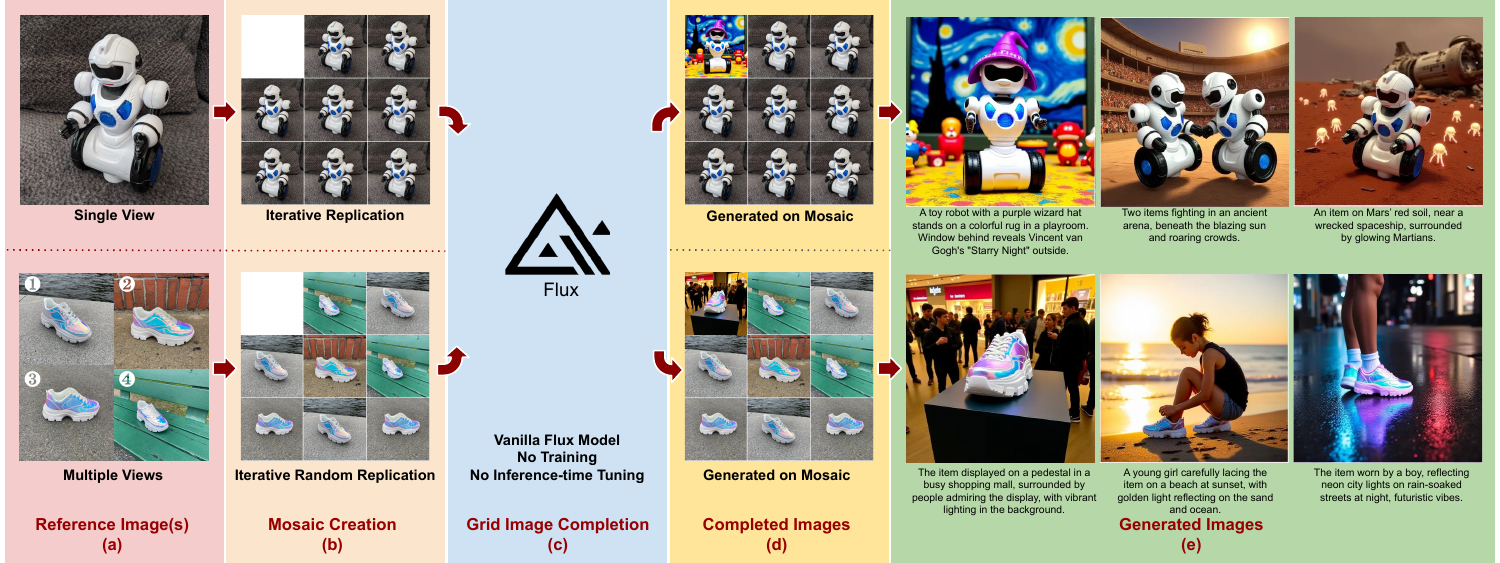}}}
    \captionof{figure}{We introduce a streamlined framework for subject-driven generation using a vanilla Flux model, with no training, no inference-time tuning, and no additional data. By leveraging mosaic-formatted image conditions and framing the task as mosaic image completion, it supports both single and multiple views, preserving subject identity and adhering to edit prompts, yielding diverse, high-fidelity results.}
    \label{fig:teaser}
\end{center}
}]
\begingroup
\renewcommand\thefootnote{}\footnotetext{*Equal contribution. 
}
\addtocounter{footnote}{-1}
\endgroup

\begin{abstract} 
We propose a simple yet effective zero-shot framework for subject-driven image generation using a vanilla Flux model. By framing the task as grid-based image completion and simply replicating the subject image(s) in a mosaic layout, we activate strong identity-preserving capabilities without any additional data, training, or inference-time fine-tuning. This “free lunch” approach is further strengthened by a novel cascade attention design and meta prompting technique, boosting fidelity and versatility. Experimental results show that our method outperforms baselines across multiple key metrics in benchmarks and human preference studies, with trade-offs in certain aspects. Additionally, it supports diverse edits, including logo insertion, virtual try-on, and subject replacement or insertion. These results demonstrate that a pre-trained foundational text-to-image model can enable high-quality, resource-efficient subject-driven generation, opening new possibilities for lightweight customization in downstream applications. 
Code: \href{https://github.com/bytedance/LatentUnfold}{https://github.com/bytedance/LatentUnfold} 
\end{abstract}
\section{Introduction}
\label{sec:intro}
Subject-driven generation~\cite{gal2022image,ruiz2023dreambooth,wei2023elite,shi2024instantbooth} focuses on producing high-quality images that accurately represent specific subjects based on provided examples or textual descriptions. This technique has a wide range of applications, including personalized content generation, advertising, marketing, artistic creation, as well as virtual reality and gaming. 

Many previous approaches utilize SD1.5~\cite{rombach2022high} and SDXL~\cite{podell2023sdxl}, built upon the UNet~\cite{ronneberger2015u} architecture, often relying on training or inference-time fine-tuning to enhance performance. More recent MMDiT~\cite{esser2024scaling,black2024flux}-based Text-to-Image (T2I) models have significantly improved generation quality. However, these advancements also result in larger models, making fine-tuning more computationally expensive and challenging. IC-LoRA~\cite{lhhuang2024iclora} trains LoRAs~\cite{hu2021lora} with task-specific data to trigger in-context generation capabilities. It showcases the ability of Flux~\cite{black2024flux} to generate identity-consistent subject views within a single image as a mosaic, utilizing specifically formatted prompts. This motivates us to consider that the base model inherently possesses knowledge about subject identity, the key is how to \textbf{activate} it.


Our hypothesis is that by fully leveraging the potential of these existing foundation models, we can avoid excessive fine-tuning and data dependency. Without reinventing the wheel, we shift our focus to inference-time computing~\cite{ilyas2024pretraining}, ensuring simplicity and resource effectiveness, and allowing the models to be deployed more seamlessly in practical applications. Concurrently, Diptych Prompting~\cite{shin2024diptych} utilizes a diptych pair format with an additional ControlNet~\cite{zhang2023adding,alimama2024control} model to perform text-conditioned inpainting for the subject in a zero-shot manner. While we aim to simplify, identifying the most minimalistic method that supports subject-driven generation, and build upon it.

Our key observation is that \textbf{a modern foundational T2I model is sufficient to enable training-free, inference-time tuning-free, subject-driven image generation}. We frame the problem as mosaic image completion, where the identity preservation capability is activated using a specially designed mosaic-formatted image combined with a proper text prompt as input, as shown in Figure~\ref{fig:teaser}. In this mosaic image, a blank section represents the desired output area, which the base model like vanilla Flux~\cite{black2024flux} fills in based on the text prompt. To complete the blank section, we use naive inpainting~\cite{avrahami2022blended} through the base mode as an efficient solution. 

Moreover, for the layout of the mosaic image, our insight is that \textbf{iterative replication} of the same input subject image can significantly improve identity preservation in the results. For example, in a $3\times3$ mosaic grid, we can designate the upper left panel as the blank section and repeatedly paste the input subject image into the remaining $8$ panels as the input condition to achieve high-fidelity results. With such a setting, this method supports not only single-view subject input with repeated mosaic panels but also multi-view input in a natural manner.

Building on these discoveries, we introduce a framework called \textbf{LatentUnfold}, which harnesses the unique mosaic subject identity preservation property in latent space and further explores enhanced attention mechanisms and prompting techniques to achieve higher fidelity.
Experiments demonstrate that our streamlined approach is sufficient to achieve state-of-the-art subject-driven generation—without requiring additional data or model fine-tuning. Our method consistently preserves subject identity while exhibiting strong adherence to edit prompts, outperforming or competing with more specialized baselines that depend on test-time tuning or domain-specific training. 

Beyond novel view generation, our LatentUnfold framework also supports diverse editing tasks, including logo insertion, virtual try-on, subject insertion, and replacement, all within a unified framework. Collectively, these results illustrate that a lightweight, inference-only method can fully exploit the advancements of modern foundation models like Flux for subject-driven image synthesis, offering a compelling and practical alternative to resource-intensive fine-tuning pipelines. Furthermore, our minimalistic approach not only simplifies subject-driven generation but also highlights broader possibilities for designing and deploying foundation models in downstream visual tasks.

\noindent In summary, we highlight the following contributions:

\vspace{0.25em}
\noindent\textbf{1. Mosaic-Driven Subject Preservation.} We reveal a straightforward yet powerful way to activate a text-to-image model’s innate object identity preservation capabilities. By arranging repeated references to the subject in a mosaic and feeding this layout into the model, we eliminate reliance on supplementary training, fine-tuning, or external data.
\vspace{0.25em}

\noindent\textbf{2. LatentUnfold Framework.} To streamline our method, we introduce an inference-time mosaicing approach that naturally accommodates both single-view and multi-view subject inputs, enabling robust zero-shot subject generation and editing.
\vspace{0.25em}

\noindent\textbf{3. Cascaded Attention Mechanism.} Leveraging insights from mosaic-based attention maps, we devise a novel multi-scale approach that pools and upscales attention across resolutions. Inserting coarse‐scale identity cues at finer scales, the cascaded attention preserves consistent subject features across the mosaic.

Building on these advancements, extensive experiments demonstrate that our approach outperforms not only methods within the same category but also those that rely on additional parameters or external data. 

\vspace{-0.5em}
\section{Related Work}
\label{sec:rw}
\vspace{-0.1em}
\subsection{Subject-driven Image Generation}
\vspace{-0.3em}
\label{sec:rw_sdig}
Subject-driven image generation preserves a subject's identity—appearance, shape, structure, and texture—while enabling modifications like view angle or context. Earlier work focused on human subjects, using GANs~\cite{goodfellow2014generative} for identity preservation in generation and editing~\cite{jian2017dual,he2019attgan,richardson2021encoding, tero2021style,shoshan2021gancontrol}, while recent diffusion models~\cite{rombach2022high,podell2023sdxl,black2024flux} have broadened this to more general subjects. One approach personalizes via test-time fine-tuning with a small set of subject images~\cite{gal2022image,ruiz2023dreambooth,liu2023cones,alaluf2023neural,voynov2023p+}, though this can be time-consuming and degrade the pretrained model's generation ability. Recent work has avoided fine-tuning by leveraging more data and task-specific designs~\cite{wei2023elite,ye2023ip,song2024imprint,jia2023taming,shi2024instantbooth,ma2024subject,patel2024lambda}, enabling zero-shot capabilities~\cite{pan2023kosmos,chen2024anydoor,li2024blip,tan2024ominicontrol,zeng2024jedi,chen2024subject,wang2024ms,zhang2024ssr,chen2024unireal}. In contrast to prior works, we focus on exploiting a single powerful Flux model~\cite{black2024flux} for subject-driven image generation—without additional data, training, or architectural changes—while achieving competitive results. 

\vspace{-0.1em}
\subsection{Image Completion}
\vspace{-0.3em}
\label{sec:rw_ic}
Framing this problem as grid image completion, inpainting~\cite{bertalmio2000inpainting,quan2024deep} emerges as a natural solution. Inpainting has evolved alongside generative models, progressing from early methods using VAEs~\cite{zheng2019pluralistic,peng2021generating} and GANs~\cite{liu2021pd,zheng2022image} to recent advances with diffusion models~\cite{lugmayr2022repaint,avrahami2023blended,zhang2023adding}, and further enhanced by text-guided approaches~\cite{avrahami2022blended,xie2023smartbrush,ju2024brushnet}, which have achieved remarkable results. While many works focus on semantic-aware methods for spatial coherence, we prioritize layout-aware methods to complete one sub-panel as a whole piece of image. Blended Latent~\cite{avrahami2022blended,avrahami2023blended} provides a simple yet effective solution, as layout attention is inherently encoded in the base model, outperforming fine-tuned models like Flux.1-Fill~\cite{black2024fill} for our task.

\vspace{-0.1em}
\subsection{Grid-based Image Generation}
\vspace{-0.3em}
\label{sec:rw_gig}
Recent T2I models~\cite{openai2023dalle3,podell2023sdxl,esser2024scaling,black2024flux} can generate identity-preserving subject views within a single $M{\times}N$ grid-based mosaic image, which contains sub-views triggered by specifically designed prompts. Some studies have used this property for generating synthetic training datasets~\cite{hui2024hq,tan2024ominicontrol}, such as OmniControl~\cite{tan2024ominicontrol}, which created 200K same-object image pairs using the Flux model~\cite{black2024flux}. IC-LoRA~\cite{lhhuang2024iclora} further enhances in-context generation by fine-tuning a LoRA~\cite{hu2021lora} with grid-based concatenated images and prompts, although it faces lower visual consistency, especially in identity transfer. ObjectMate~\cite{winter2024objectmate} trained an SDXL~\cite{podell2023sdxl} styled model using a $2{\times}2$ mosaic, leveraging recurrence priors for subject-driven image generation. Unlike these methods, our approach doesn't rely on additional data or training. We also acknowledge Diptych Prompting~\cite{shin2024diptych}, which treats the problem as an inpainting task for a $1{\times}2$ diptych image pair, achieving good results under zero-shot settings with attention enhancement. However, their method depends on a ControlNet model~\cite{alimama2024control}, while ours uses only a single base model for comparable results, improving identity preservation through iterative replication without modifying attention values or adding hyperparameters, and naturally supporting multiview reference images.

\vspace{-0.5em}
\section{Method}
\begin{figure*}
    \makebox[\textwidth]{\includegraphics[width=0.9\textwidth, trim=0 140 0 0, clip]{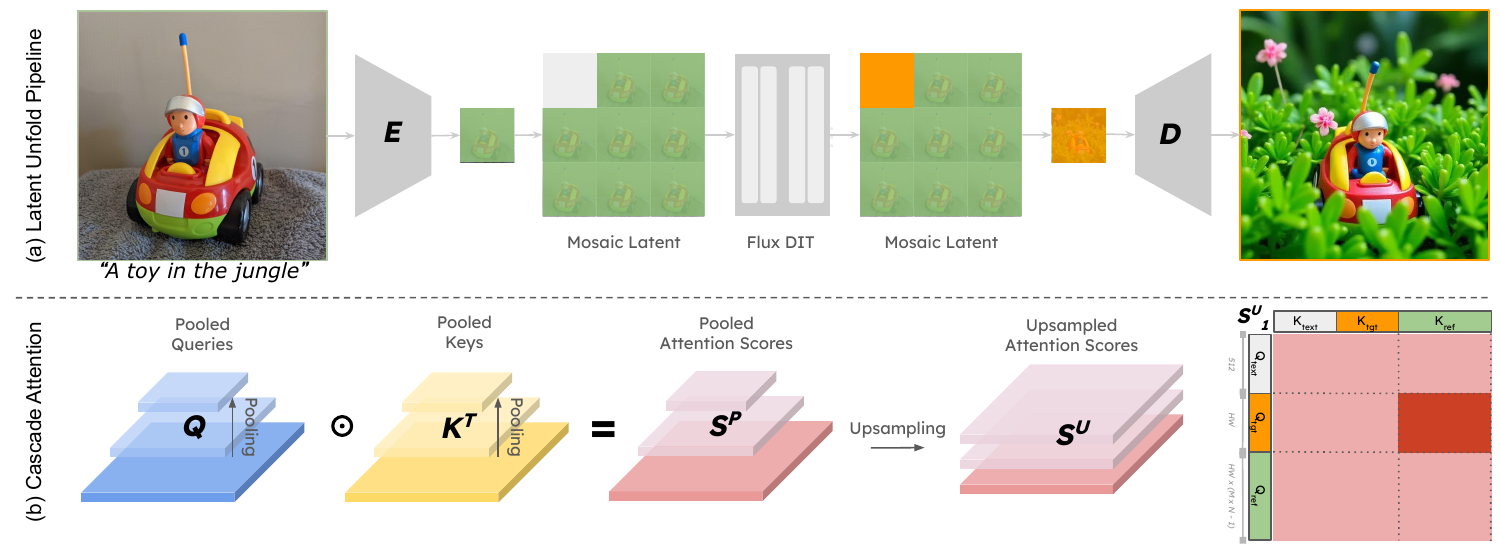}}
    \vspace{-2.5em}
    \caption{The top row illustrates the LatentUnfold pipeline, while the bottom row showcases the novel Cascade Attention mechanism.}
    \vspace{-1.2em}
    \label{fig:method_pipeline}
\end{figure*}

\vspace{-0.5em}
We observe that a \textit{mosaic-formatted} image with iterative replication of the subject image across the grid yields satisfied results in subject-driven image generation using a single T2I model. To further streamline the construction of conditions, we propose a framework called \textbf{LatentUnfold}, which operates in the latent space, as shown in the top row of Figure~\ref{fig:method_pipeline}. The details of LatentUnfold and mosaic image completion are provided in Sections~\ref{sec:method:latent_unfold} and~\ref{sec:method:mosaic_completion}. To further exploit the properties of the mosaic approach, we introduce a novel attention enhancement method called \textbf{Cascade Attention} to achieve higher-quality generation. This method is illustrated in Figure~\ref{fig:method_pipeline} and explained in detail in Section~\ref{sec:method:cascade_attention}. Finally, we present \textbf{Meta Prompting} as an additional option, which enables Multimodal Large Language Model (MLLM) to generate improved prompts, seamlessly integrating with LatentUnfold to fully activate subject identity preservation. This technique is detailed in Section~\ref{sec:method:meta_prompting}.

\label{sec:method}
\vspace{-0.3em}
\subsection{Latent Unfold}
\vspace{-0.3em}
\label{sec:method:latent_unfold}
Our method constructs a \textit{mosaic-formatted} $M{\times}N$ grid (e.g., $3\times3$) for zero-shot subject-driven generation, with one blank panel as the target area and the rest filled with repeated subject images (Figure~\ref{fig:teaser}b). For clarity, we refer to each rectangular region within the mosaic (the tiles) as a \textit{panel}. This design enhances identity consistency during generation. While this process can occur in pixel space with additional preprocessing, shifting it to latent space offers a more elegant and robust solution.

We implement this by unfolding the reference image latent into a mosaic latent. As shown in Figure~\ref{fig:method_pipeline}a, the reference image is first encoded into a latent code $\bL_{r}$, which is then tiled into an $M{\times}N$ \textit{mosaic} latent $\bL$, where the upper-left panel is just a placeholder filled with zeros:
\vspace{-0.5em}
\begin{equation}
\bL \;=\;
\begin{pmatrix}
\mathbf{0} & \bL_r & \cdots & \bL_r \\
\vdots & \vdots & \ddots & \vdots \\
\bL_r & \bL_r & \cdots & \bL_r
\end{pmatrix}  \in \mathbb{R}^{M\cdot H \times N\cdot W \times C},
  \vspace{-0.4em}
\end{equation}

\noindent where $H,W,C$ are the height, width and channels of the latent respectively. Decoding $\bL$ produces a mosaic image with $M\cdot N-1$ reference images and an empty top-left panel.
 
The LatentUnfold is highly flexible, supporting both single-view (Figure~\ref{fig:teaser} top) and multi-view (Figure~\ref{fig:teaser} bottom) input subject images. For multi-view inputs, the mosaic latent can incorporate different perspectives of the subject latents through iterative random replication, enabling diverse yet consistent outputs.

\vspace{-0.3em}
\subsection{Mosaic Image Completion}
\vspace{-0.3em}
\label{sec:method:mosaic_completion}

Our goal is to \emph{edit} a specific panel (e.g., the top-left) within a mosaic image (composed of multiple panels), while preserving the other panels that contain reference images of our subject. We build on a denoiser-based \emph{completion} pipeline, adapting standard inpainting ideas~\cite{avrahami2022blended} so that only a chosen panel is updated, while the rest remains untouched. The procedure is summarized in \Cref{alg:mosaic_completion} and described below.

\noindent\textbf{Setup.}
Let $\bL$ denote the mosaic image arranged in an $M \times N$, as detailed in Section~\ref{sec:method:latent_unfold}. We define a binary mask $\bm \in \{0,1\}^{M\cdot H \times N\cdot W}$ that identifies the panel to be generated (i.e. the top-left panel) with $\bm=1$, and $\bm=0$ elsewhere. We sample an initial noise latent $\bx_0 \sim \mathcal{N}(\mathbf{0}, \mathbf{I})$ and denote $T$ as the number of denoising steps.

\noindent\textbf{Noise Level Consistency.}
A crucial detail is that the denoiser model expects a uniform noise level across the entire image at each step. However, we need the unmasked (reference) panels of $\bL$ to remain intact throughout the process. To reconcile these needs, we blend noise into $\bL$ following the noise schedule used to train the base FLUX model we use in our experiments ~\cite{black2024flux, lipman2023flowmatching}. We define a discrete schedule by letting $
  t \;=\; \frac{i}{T-1},
  \quad i \in \{0,1,\dots,T-1\}.
$
At time step \(t\), we form a “noised” version of the mosaic:
\vspace{-0.5em}
\begin{equation}
  \bL_t \;=\; (1 - t)\,\bx_0 
  \;+\; t\bL.
    \vspace{-0.5em}
\end{equation}

\noindent\textbf{Masking.}
To update only the target panel, we selectively combine the newly generated content in that panel with the noised mosaic in the rest of the image. At time $t$, given the partially denoised latent $\bx_t$, we construct:
\vspace{-0.5em}
\begin{equation}
  {\bxt}_t \;=\; \bm \,\odot\, \bx_t \;+\; \bigl(\mathbf{1} - \bm\bigr)\,\odot\, \bL_t.
  \vspace{-0.5em}
\end{equation}
Hence, inside the masked region ($\bm=1$), we use the latest $\bx_t$ generated by the model, while outside that region ($\bm=0$), we revert to the “noised” mosaic $\bL_t$. This arrangement ensures that only the target panel is being synthesized, while the unmasked panels remain unmodified (up to the noise level blending that keeps them consistent with the denoiser’s expectations).

\noindent\textbf{Denoising.}
We then feed $\widehat{\bx}_t$, along with the text prompt $\mathbf{p}$, into a text-to-image denoiser $\mathcal{D}$:
\vspace{-0.5em}
\begin{equation}
  \bx_{t+\Delta t} \;=\; \mathcal{D}\bigl(\widehat{\bx}_t,\, \mathbf{p},\, t\bigr),
    \vspace{-0.5em}
\end{equation}
where $\Delta t = \tfrac{1}{T}$. After the $T$ steps, we decode $\bx_T$ to the pixel space, yielding the mosaic image whose target panel is newly generated according to the text prompt, while the other panels remain faithful to the original mosaic (i.e., containing the subject reference).

\noindent\textbf{Integration with Editing.}
Due to the flexible LatentUnfold framework, inpainting process can seamlessly introduce \emph{new} content into the unmasked panel. Meanwhile, the repeated subject references in the masked panels that are left intact, help the denoising model maintain the subject's identity. After $T$ steps, the final output $\bx_T$ will thus \emph{edit} the designated region according to user instructions while preserving the original references of the subject.

\begin{algorithm}[t]
\caption{Mosaic Completion}
\label{alg:mosaic_completion}
\begin{algorithmic}[1]
\REQUIRE Mosaic $\bL \in \mathbb{R}^{(M\cdot H) \times (N\cdot W) \times C}$,mask $\bm \in \{0,1\}^{(M H)\times(N W)}$,denoiser $\mathcal{D}$,prompt $\mathbf{p}$, steps $T$ 
\STATE Sample $\bx_0 \sim \mathcal{N}(\mathbf{0}, \mathbf{I})$
\FOR{$i = 0 \dots T-1$}
    \STATE $t \leftarrow \frac{i}{T-1}$
    \STATE $\bL_t \leftarrow (1-t)\,\bx_0 + t\,\bL$
        \COMMENT{Adapt noise level}
    \STATE $\widehat{\bx}_t \leftarrow \bm\,\odot\,\bx_{t} \;+\; (\mathbf{1}-\bm)\,\odot\, \bL_t$
        \COMMENT{Apply mask}
    \STATE $\bx_{t+\Delta t} \leftarrow \mathcal{D}(\widehat{\bx}_t, \mathbf{p}, t)$
        \COMMENT{Denoise}
\ENDFOR
\RETURN $\bx_T$
\end{algorithmic}

\end{algorithm}

\vspace{-0.3em}
\subsection{Cascade Attention}
\begin{figure}
    \centering
    \includegraphics[width=\columnwidth, trim=0 50 342 0, clip]{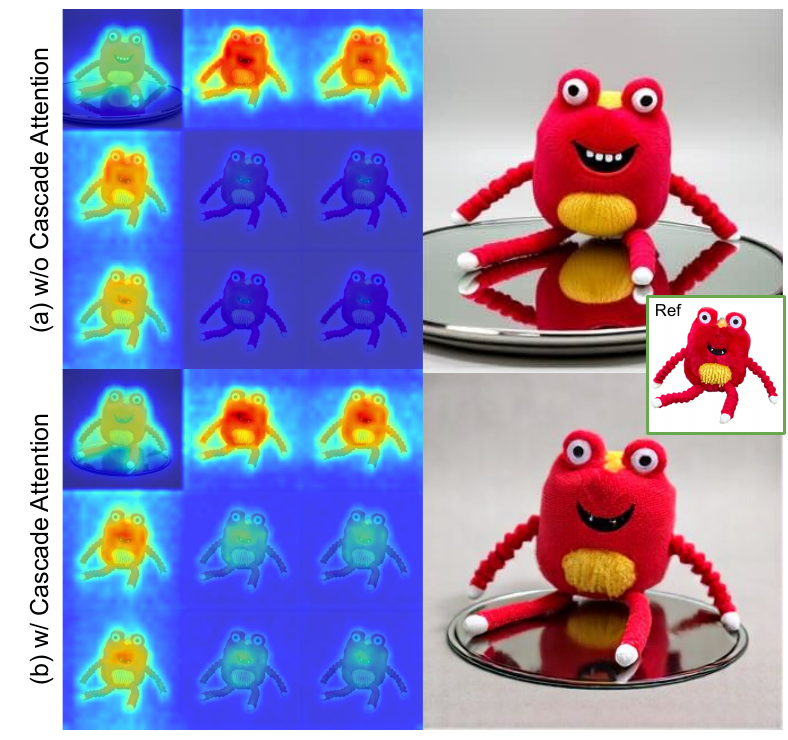}
    \caption{The attention visualizations, both with and without Cascade Attention, are shown. The reference image is displayed on the right. Notice the enhanced attention, leading to better detail preservation on the toy's teeth and belly.}
    \label{fig:cascade_attention}
\vspace{-1.5em}
\end{figure}

\vspace{-0.3em}
We visualize the attention mechanism of mosaic image completion \cite{attention_map_diffusers} in Figure~\ref{fig:cascade_attention}a. We observe that the model attends to the reference subject across different panels, and there are variation of the attention patterns due to positional encoding. We observe though that not all panels are attended equally and the model misses some texture details.  Building on this observation, we design a \textbf{Cascade Attention}. This provides an iterative mechanism that looks a the reference subject at different scales as illustrated in Figure~\ref{fig:cascade_attention}. We begin with the standard attention setup but instead of relying solely on a single fine‐scale representation, we also construct pooled (downsampled) versions of the queries and keys to capture a more global view of the subject. These pooled attention score maps are then upsampled and added back to the original fine‐scale score map. This repeated “coarse‐to‐fine” feedback helps maintain a consistent subject identity while refining details across all panels.

\vspace{0.5em}
\noindent\textbf{Pooled attention maps.} 
We assume one head for simplicity. We denote the original (fine) queries and keys: 
\begin{equation}
  \mathbf{Q}_1 \in \mathbb{R}^{n \times d}, 
  \quad
  \mathbf{K}_1 \in \mathbb{R}^{n \times d},
\end{equation}
where the query positions $n=M\cdot \tfrac{H}{p}\cdot N \cdot \tfrac{W}{p}$ where $p$ is the attention patch size and $d$ the hidden dimension. 

To incorporate broader context, we define pooled queries and keys $\{\mathbf{Q}_i, \mathbf{K}_i\}$, for $i=2,\dots,I$, by average‐pooling along the spatial dimension:
\begin{equation}
  \mathbf{Q}_i = \mathrm{pool}(\mathbf{Q}_{i-1}), \quad 
  \mathbf{K}_i = \mathrm{pool}(\mathbf{K}_{i-1}).
\end{equation}

We construct the attention score maps for each layer:

\begin{equation}
  \mathbf{S}_i^P \;=\;  \mathbf{Q}_i \,\mathbf{K}_i^\top
  \;\;\in\;\; \mathbb{R}^{\frac{n}{i} \times \frac{n}{i}}.
\end{equation}

\noindent\textbf{Upsampling and Aggregation.}
Each pooled score map $\mathbf{S}_i^P, i\ge2$ has shape $\frac{n}{i} \times \frac{n}{i}$. We first \emph{bilinearly upsample} them to $S_i^U, i\ge2$ of size $n \times n$. Then we add all the upsampled maps into the original fine‐scale map to get the cascaded attention map:
\begin{equation}
  \mathbf{S}
  \;=\;
  \mathbf{S}_1^U
  \;+\;
  \sum_{i=2}^{I}
    \Bigl(\mathbf{S}_i^U\Bigl[
        Q_{\mathrm{tgt}},K_{\mathrm{ref}}\Bigr]\Bigr),
\end{equation}
where $\Bigl[Q_{\mathrm{tgt}},K_{\mathrm{ref}}\Bigr]$ indicates we only update the slice of the attention map corresponding to the target queries and reference keys (see \Cref{fig:method_pipeline}). In terms of time complexity is method only introduces a constant cost $\bigl(\sum_{i=1}^{I}\ \tfrac{1}{i^2}\bigr)O(n^2 d)$.

\vspace{0.5em}
\noindent
\textbf{Why It Helps.}  
The model uses a coarse‐scale “big picture” of the subject, then reinserts these clues at fine resolution to maintain consistent identity across multiple panels. Differences in positional encodings between $\mathbf{Q}_1,\mathbf{K}_1$ and their pooled versions introduce slight variations, helping preserve subtle details (e.g., the subject’s facial or body features). The result is a stronger alignment between reference panels and the newly generated panel, ultimately delivering sharper, more faithful details compared to a single‐scale approach.

\label{sec:method:cascade_attention}

\subsection{Meta Prompting}
\label{sec:method:meta_prompting}

Modern T2I models can produce very high-quality and detailed outputs. Nevertheless, prompting them in an effective way is a laborious tasks. 
For instance, the toy car depicted in Fig.~\ref{fig:method_pipeline}(a) can be generated using the following formatted prompt with LatentUnfold: \textit{“This collection of full-frame images showcases a colorful toy, consistently positioned within a realistic environment. [IMAGE1] presents the toy in a jungle setting, while [IMAGE2] through [IMAGE9] emphasize its vivid colors and playful design.”}

Using MLLMs to refine initial prompts significantly enhances results and enables more precise editing through detailed, well-crafted prompts \cite{hao2023optimizingpromptstexttoimagegeneration, Wang_2024, rout2024semanticimageinversionediting, lhhuang2024iclora}. We build on this by introducing a method where an MLLM, given a \textit{meta-prompt} and a reference image, generates editing prompts for a T2I model. In our case, we use GPT-4o as the MLLM and FLUX as the T2I model, though the approach is model-agnostic. Examples of meta-prompts are shown in Box~\ref{box:meta_prompt_v1}.

\begin{floatcolorbox}[box:meta_prompt_v1]{Meta-Prompt}{green}  
\tiny

When given an image, you make a mosaic image consisting of a 3x3 grid of sub-images showing the exact same subject, describe each subject's appearance in sub-images sequentially from top-left to bottom-right. Limit each description to 50 words. Describe details especially unique appearance like logos, colors, textures, shape, structure and material that can recreate the subject. Refrain from any speculative or guesswork. 
\textbf{Output Format Example:}\\
\texttt{\detokenize{
{
    "row1": {
        "image1": "highlights the sneaker's white laces and textured sole, emphasizing its casual style.",
        "image2": "captures the sneaker's unique color combination and material texture from a slightly angled view.",
        "image3": "displays a close-up of the sneaker's mint green and lavender panels, focusing on the stitching details."
    },
    "row2": {
        "image1": "presents the sneaker's side, showing the yellow stripe and layered design elements.",
        "image2": "showcases the sneaker's rounded toe and smooth material finish, highlighting its modern aesthetic.",
        "image3": "features the sneaker's interior lining and padded collar, emphasizing comfort and design."
    },
    "row3": {
        "image1": "focuses on the sneaker's sole pattern and grip, showcasing its practical features.",
        "image2": "captures the sneaker's overall shape and color scheme, providing a comprehensive view.",
        "image3": "features the structure and texture of the subject."
    },
    "summary": "This set of full-frame photos captures an identical pastel-colored sneaker subject firmly positioned in the real scene, highlighting its unique design, color scheme, and material details from various perspectives (cinematic, epic, 4K, high quality)."
}
}}\\
For content in summary: it should starts with This set of full-frame photos captures an identical xxx subject and include firmly positioned in the real scene.
\end{floatcolorbox}
\section{Experiments}
\label{sec:exp}
\label{sec:imp}

\subsection{Benchmark Details.}

We evaluate our method using the DreamBooth benchmark~\cite{ruiz2023dreambooth}, a standard for assessing editing approaches~\cite{shin2024diptych, ye2023ip, patel2024lambda, pan2023kosmos, li2024blip, fei2023gradient}, which includes 30 subjects across 15 classes (9 live subjects, such as dogs and cats, and 21 inanimate objects). Each subject is captured from 4 to 6 views under varying conditions and paired with 25 prompts tailored to the subject’s type. We expand on the minimal editing instructions provided by DreamBooth (e.g., "a backpack in the jungle") by incorporating meta prompting. We measure generation quality using identity preservation and text alignment metrics: CLIP-I~\cite{radford2021learning-clip} and DINO~\cite{caron2021emerging-dino} for identity fidelity, assessing how well the generated image retains the subject’s features and structure, and CLIP-T~\cite{radford2021learning-clip} for text alignment, which measures how accurately the image matches the prompt. 

\subsection{Baseline Comparisons}
\label{sec:baseline}
\noindent\textbf{Quantitative Results}
\label{sec:qualitative}
We considered $3\times3$ grid configurations, removing the background~\cite{BiRefNet} from the reference image, while generating output images at a resolution of $512\times512$. We used GPT-4o or the meta-prompting. For the T2I synthesis, we utilized FLUX.1-dev~\cite{black2024flux}. Following the original method, we performed $T=28$ editing steps. Our initial experiments showed that stronger guidance ($g=7.0$) improved the quality of edits, and thus we maintained this setting in all subsequent experiments. We
tested on 750 text-reference pairs ($25$ prompts~$\times$ $30$ subjects)
from DreamBooth dataset~\cite{ruiz2023dreambooth} with 4 different seeds.

We compare our approach with three main categories of subject-driven generation methods:(1) \textbf{Extra Params} methods, (2) \textbf{Extra Data} methods, and (3) \textbf{All-free} methods, which our method falls into. While methods in the first two groups generally do not require fine-tuning at test time, they still depend on domain-specific training phases, which can be resource-intensive and often necessitate the collection of high-quality data— a challenging requirement to scale.

\begin{table}[t]
\centering
\caption{Comparison of related methods with their respective metrics. IP-Adapter$^a$ was evaluated in~\cite{shin2024diptych}, Diptych$^b$ is re-implemented in this work. LatentUnfold\textsuperscript{avg} denotes the average CLIP-I and DINO scores across the three reference images, while LatentUnfold\textsuperscript{best} selects the highest score among them. The scores are shown in percentage.}
\label{tab:res:baseline_comparison_main}
\resizebox{\columnwidth}{!}{%
\begin{tabular}{clccccc}
\toprule
\textbf{Type} & \textbf{Method} & \textbf{Base} & \textbf{Params} & \textbf{CLIP-I}$\uparrow$ & \textbf{DINO}$\uparrow$ & \textbf{CLIP-T}$\uparrow$ \\
\midrule
\multirow{3}{*}{\shortstack{\shortstack{Extra\\ Params}}}
& IP-Adapter$^a$\small{\cite{ye2023ip}}      & FLUX & 22M & 72.5 & 56.1 & \textbf{35.1} \\
& OminiControl \small{\cite{tan2024ominicontrol}}              & FLUX   & 15M    & 77.3 & 62.7 & 32.2 \\
& Diptych$^b$~\cite{shin2024diptych} & FLUX & 2B & \textbf{79.4} & \textbf{66.6} & 31.9 \\
\midrule
Extra Data & Re-Imagen \cite{chen2022re}          & Imagen & -- & 74.0 & 60.0 & 27.0 \\
\midrule
\multirow{2}{*}{\shortstack{All-free\\1 View}}
& RF-Inversion \small{\cite{rout2024semanticimageinversionediting}} & FLUX & -- & \textbf{78.7} & 61.9 & 29.4 \\
& LatentUnfold & FLUX & --  & \textbf{78.7} & \textbf{64.0} & \textbf{30.5} \\
\midrule
\multirow{2}{*}{\shortstack{All-free\\3 Views}}
& LatentUnfold\textsuperscript{avg} & FLUX & -- & 77.6 & 61.8 & \textbf{30.5} \\
& LatentUnfold\textsuperscript{best} & FLUX & -- & \textbf{80.6} & \textbf{66.0} & \textbf{30.5} \\
\bottomrule
\vspace{-2.5em}
\end{tabular}}
\end{table}

\Cref{tab:res:baseline_comparison_main} compares our method with others in terms of identity preservation and text alignment.
Despite not having any dedicated training or specialized data collection, our approach achieves competitive performance compared to methods in the other two groups and outperforms methods within the same group. These findings highlight the practical advantages of our method enable flexible, high-fidelity subject editing without the computational overhead of test-time tuning or the data-collection constraints of training-based approaches. 

\noindent\textbf{Human Preference Study}
As noted by~\cite{ruiz2023dreambooth}, numerical metrics have limitations. Similarly, we observe a notable discrepancy. For example, in Figure~\ref{fig:cascade_attention}, while the bottom row clearly exhibits higher identity similarity to the reference image than the top, the calculated scores (subject only, without background) show the bottom row with a CLIP-I score of $0.937$ and a DINO score of $0.797$, compared to the top row's CLIP-I score of $0.938$ and DINO score of $0.806$, which contradicts human perception.

To address this, we conducted a human preference study, which further evaluates and demonstrates that our method outperforms two state-of-the-art approaches most relevant to ours: the pre-trained OmniControl~\cite{tan2024ominicontrol} and the inpainting model-dependent Diptych~\cite{shin2024diptych}. In a pairwise manner, we collected $1500$ responses from $15$ participants, with $750$ responses for each method compared to ours. The study considers three key dimensions: identity preservation, text alignment, and image quality, under a single-view setting. As shown in Table~\ref{tab:user}, our method consistently outperforms the compared methods in identity preservation and image quality while surpassing Diptych and performing comparably to OmniControl in text alignment.

Regarding identity preservation, we believe mosaic condition and Cascade Attention mechanism improves performance, as shown in Figure~\ref{fig:cascade_attention}, by effectively transferring subject details that enhance human perception. In terms of image quality, the additional parameters in the other two methods degrade the base model's output.


\begin{table*}[t]
\centering
\caption{Human preference comparison of methods based on Subject Identity, Text Alignment, and Image Quality. Diptych$^a$ is re-implemented in this work.}
\label{tab:user}
\resizebox{0.8\textwidth}{!}{%
\begin{tabular}{c|ccc|ccc|ccc}
\toprule
\textbf{Method} & \multicolumn{3}{c|}{\textbf{Subject Identity \%}} & \multicolumn{3}{c|}{\textbf{Text Alignment \%}} & \multicolumn{3}{c}{\textbf{Image Quality \%}} \\
& \textbf{Win} & \textbf{Lose} & \textbf{Tie} & \textbf{Win} & \textbf{Lose} & \textbf{Tie} & \textbf{Win} & \textbf{Lose} & \textbf{Tie} \\
\midrule
LatentUnfold vs Diptych$^a$ \cite{shin2024diptych} & 41.87 & 30.80 & 27.33 & 22.00 & 20.93 & 57.07 & 45.33 & 31.34 & 23.33 \\
LatentUnfold vs OminiControl \cite{tan2024ominicontrol} & 44.53 & 26.27 & 29.20 & 18.40 & 23.33 & 58.27 & 31.47 & 30.66 & 37.87 \\
\bottomrule
\end{tabular}}
\end{table*}

\subsection{Ablation Studies}
\label{sec:ablation}

We conduct a series of ablations to evaluate how different design choices affect the performance of our approach. Most settings are carried over from our main qualitative experiments in Section~\ref{sec:qualitative}. However, unlike the main experiments, we use a single seed across 750 text-reference pairs ($25$ prompts~$\times$ $30$ subjects) for each ablation study. Specifically, we investigate the effect of (i) mosaic grid shape, (ii) background removal, (iii) cascade attention, and (iv) meta prompting on a single view of the subject.

The results, summarized in Table~\ref{tab:res:ablation}, are reported in terms of identity preservation (CLIP-I, DINO) and text alignment (CLIP-T). Overall, these ablations confirm that the $3\times3$ grid tends to yield better performance than smaller grids. Background removal significantly enhances image-text alignment. Cascade Attention design positively impacts subject identity preservation by a notable margin. Meanwhile, Meta Prompting further strengthens subject identity preservation but slightly compromises image-text alignment. These findings suggest that practitioners can fine-tune these design choices based on their priorities.

\begin{table}[ht]
\centering
\vspace{1em}
\caption{Ablations of our method with different settings and components. The scores are shown in percentage.}
\resizebox{\columnwidth}{!}{%
\begin{tabular}{ccccccc}
\toprule
\textbf{Grid} & \textbf{Segment} & \textbf{Cascade} & \textbf{Prompt} & \textbf{CLIP-I}$\uparrow$ & \textbf{DINO}$\uparrow$ & \textbf{CLIP-T}$\uparrow$  \\
\midrule
\multirow{1}{*}{1x2}
  & \multirow{1}{*}{ }
  &   &   & 69.9 & 42.6 & 30.8  \\
\midrule
\multirow{1}{*}{2x2}
  & \multirow{1}{*}{ }
  &   &   & 71.1 & 47.0 & 30.9 \\
\midrule
\multirow{1}{*}{3x3}
  & \multirow{1}{*}{ }
  &   &   & 74.2 & 54.1 & 29.0 \\
\midrule
\multirow{1}{*}{3x3}
  & \multirow{1}{*}{ \True}
  &   &   & 73.9 & 59.0 & 30.7 \\
\midrule
\multirow{1}{*}{3x3}
  & \multirow{1}{*}{ \True}
  & \True  &   & 75.1 & 60.2 & \textbf{31.0} \\
\midrule
\multirow{1}{*}{3x3}
  & \multirow{1}{*}{ \True}
  & \True  & \True  & \textbf{80.0} & \textbf{66.5} & 30.6 \\
\bottomrule
\end{tabular}
}
\label{tab:res:ablation}
\end{table}

\subsection{Applications}
\label{sec:applications}
In this section, we qualitatively demonstrate several potential applications of our approach, including subject-driven generation, logo insertion, virtual try-on, subject replacement, and subject insertion. Please refer to Supplemental Materials for more results.

\begin{figure*}[th]
  \centering
  \vspace{0.2em}
  \makebox[\textwidth]{\includegraphics[width=0.95\textwidth, trim=0 105 0 0, clip]{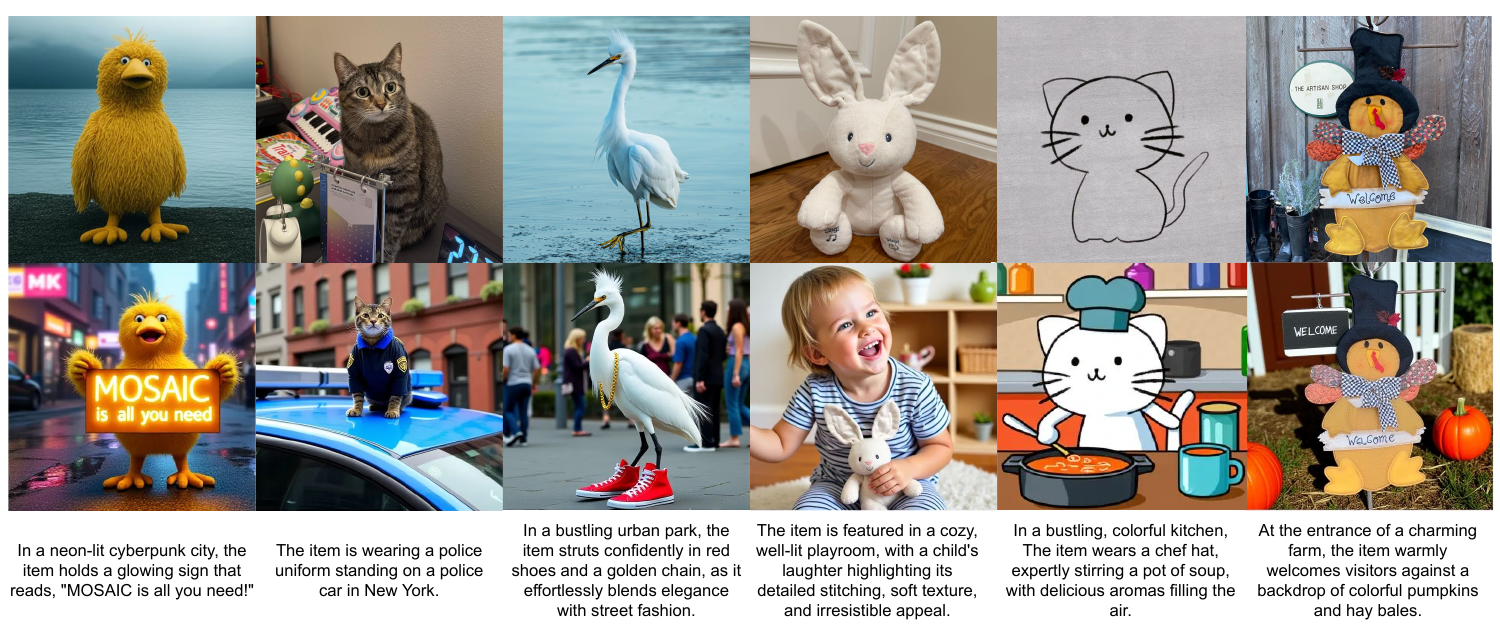}}
  \caption{Qualitative results of subject-driven tasks on novel objects. The top row displays reference images, while the corresponding text prompts are listed below each generated image.}
  \label{fig:novel}
\vspace{-2em}
\end{figure*}

\noindent\textbf{Subject-Driven Generation}
As discussed earlier in this section, our method performs well on the DreamBooth dataset~\cite{ruiz2023dreambooth}. To further evaluate its robustness, we present subject-driven task results on a wider range of novel objects, accommodating diverse inputs and controls, as shown in Figure~\ref{fig:novel}. Our method effectively preserves the subject's appearance, including structure, color, and texture, while accurately following text prompts across various subjects. However, for less common subjects with complex compound structures, the generation quality tends to degrade, as seen in the rightmost example.

\begin{figure*}[th]
  \centering
  \makebox[\textwidth]{\includegraphics[width=0.95\textwidth, trim=0 78 0 0, clip]{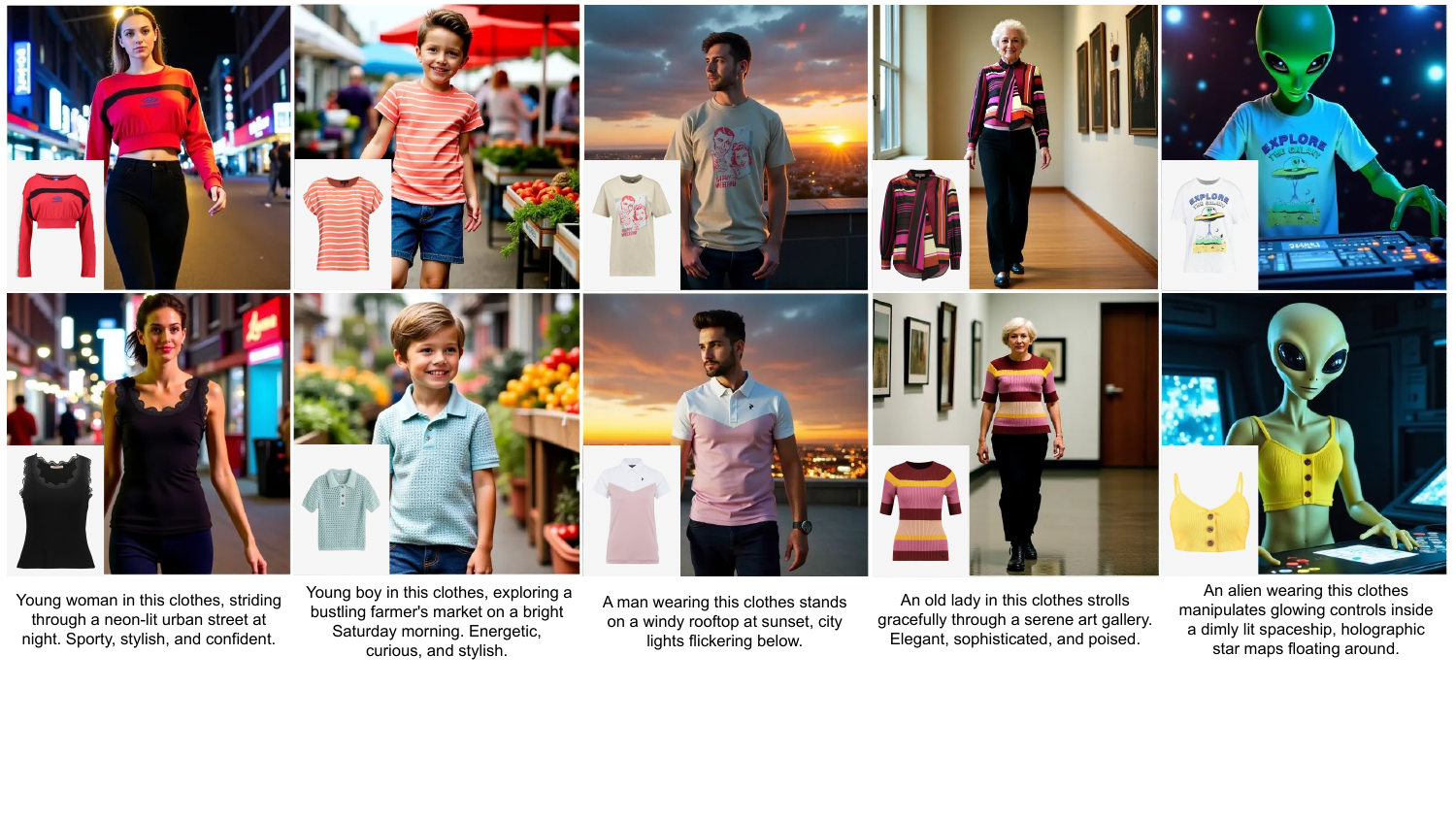}}
  \caption{Our method demonstrates robust performance in the virtual try-on task, where both the human model and the scene are generated through the control of text prompts. The reference garments are displayed as small images in the bottom-left corner of each image, and the corresponding text prompts are listed below each generated image.}
  \label{fig:vto}
\end{figure*}

\noindent\textbf{Virtual Try-on} Our method demonstrates robust performance in the virtual try-on task, effectively preserving the intricate details and textures of garments\footnote{The test images are sourced from~\cite{choi2021viton}.}. It achieves remarkable realism and consistency in rendering diverse clothing items across various body types and poses, as shown in Figure~\ref{fig:vto}. Furthermore, our method does not require an explicitly defined human model as input, instead generating a highly diverse and high-quality human model that seamlessly fits into aesthetically pleasing scenes, making it ready to use in downstream scenarios.

\begin{figure}[th]
  \centering
  \makebox[\columnwidth]{\includegraphics[width=1.0\columnwidth, trim=0 12 0 0, clip]{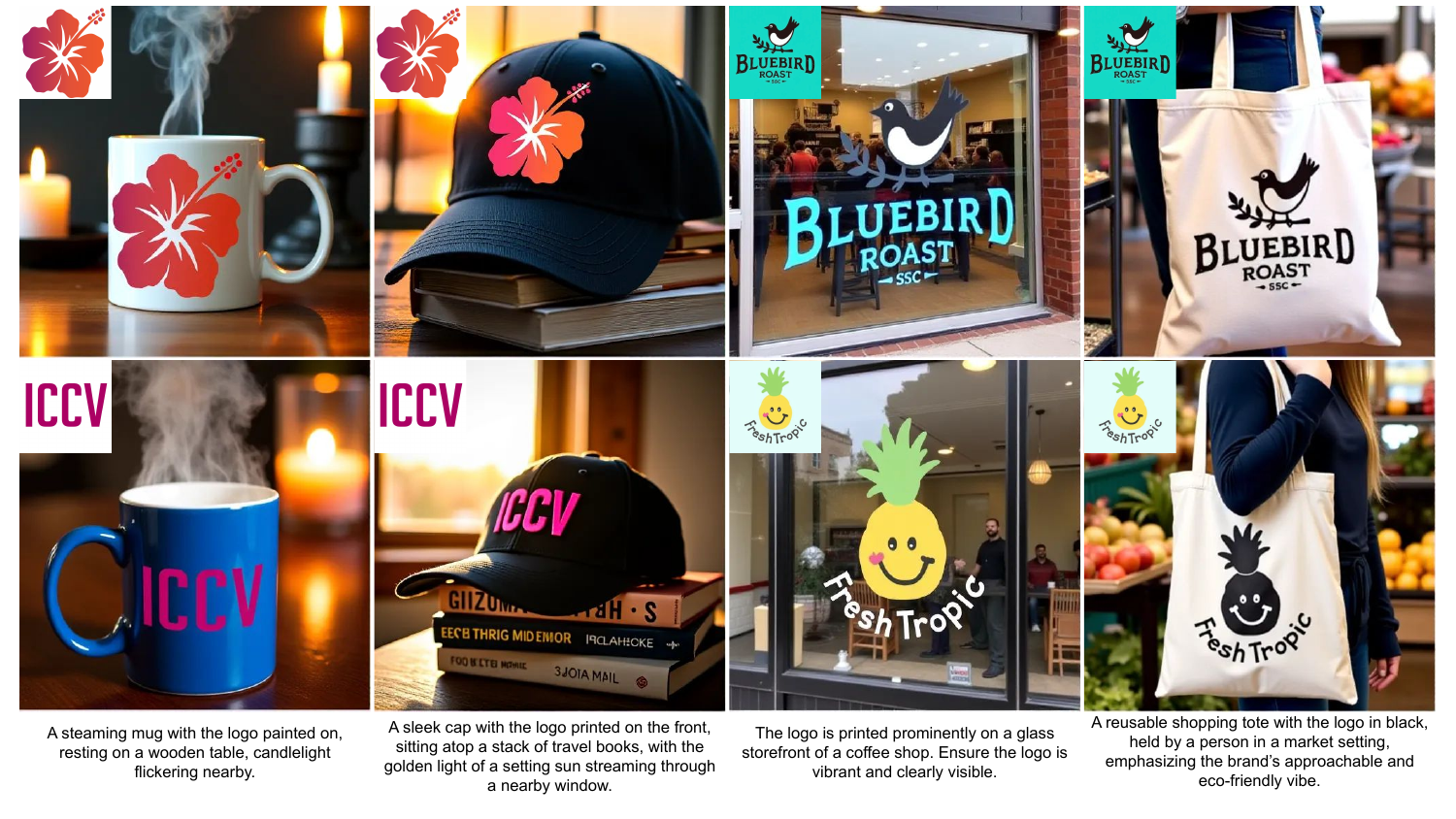}}
  \caption{Our method demonstrates strong performance in the logo insertion task. The reference
logos are displayed as small images in the upper-left corner of
each image, and the corresponding text
prompts are listed below each generated image.}
  \label{fig:logo}
  \vspace{-1.0em}
\end{figure}

\noindent\textbf{Logo Insertion} Logo insertion, gaining increasing attention from researchers~\cite{zhu2024logosticker}, has numerous use scenarios, including brand integration in marketing, personalized merchandise design, and augmented reality. Our method demonstrates strong performance in the logo insertion task, effectively preserving the unique patterns of logos\footnote{The logos are sourced from \url{iccv.thecvf.com} and~\cite{lhhuang2024iclora}.}. It achieves exceptional fidelity and coherence in generating diverse logos across various contexts, as illustrated in Figure~\ref{fig:logo}.

\begin{figure}[ht]
  \vspace{-0.4em}
  \centering
  \makebox[\columnwidth]{\includegraphics[width=0.97\columnwidth, trim=0 0 390 0, clip]{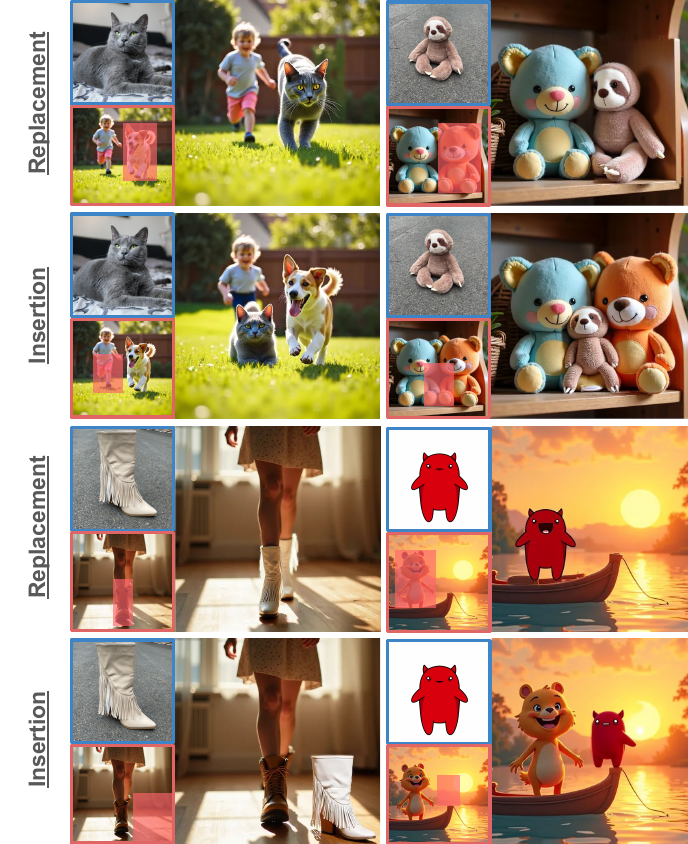}}
  \caption{Our method supports subject replacement (top) and subject insertion (bottom) tasks, with four test cases for each. The smaller image on the left with a blue border represents the reference subject, while the red-bordered image on the right represents the target to be edited. The red rectangle highlights the region to be edited, and the larger image on the right shows the final result.}
  \label{fig:edit}
\end{figure}

\noindent\textbf{Subject Replacement and Insertion} Our method supports simple subject driven image editing tasks, including subject replacement and insertion, provided a specific region is designated, as illustrated in Figure~\ref{fig:edit}. We observe that the outcomes are sensitive to the size and placement of the region, with better results typically achieved when the region is centrally located and relatively large. 

\section{Conclusion}
\label{sec:concl}
\vspace{-0.5em}
We have shown that modern T2I models like Flux respond favorably to mosaic-formatted latents, where repeated patterns significantly enhance subject-driven generation. Building on this insight, we introduced a minimalistic, training-free, and data-free approach that reframes the task as mosaic-conditioned image completion. By employing a novel cascaded attention mechanism and meta prompting, our (inference-time) method achieves high-fidelity outputs without requiring additional data or model updates.

Beyond novel view synthesis, our framework supports a range of editing tasks, including logo insertion, style transfer, subject replacement and insertion. While its reliance on mosaic layouts may affect generation speed and output resolution, these drawbacks can be addressed through improved computational resources and incremental refinements. Overall, this work demonstrates the potential of foundational models for efficient and scalable subject-driven generation, offering practical insights for future research and applications. Crucially, our method relies on the minimal assumption that attention serves as the model’s core image modeling mechanism, making it potentially model-agnostic. As more powerful image-generation architectures emerge, it would be intriguing to see how seamlessly this approach can be adapted, especially given that it requires no additional data or training.
\clearpage
\newpage
\clearpage
\section{Acknowledgments}
We sincerely acknowledge the insightful discussions from Yiding Yang, Bo Liu, Xiao Yang and Xiaohui Shen.
We genuinely thank all the participants in the user preference study, as well as Jincheng Liang, Lu Guo, and Neng Li for their valuable assistance.

\bibliographystyle{ieeenat_fullname}
\bibliography{references}

\begin{thebibliography}{71}
\providecommand{\natexlab}[1]{#1}
\providecommand{\url}[1]{\texttt{#1}}
\expandafter\ifx\csname urlstyle\endcsname\relax
  \providecommand{\doi}[1]{doi: #1}\else
  \providecommand{\doi}{doi: \begingroup \urlstyle{rm}\Url}\fi

\bibitem[Alaluf et~al.(2023)Alaluf, Richardson, Metzer, and Cohen-Or]{alaluf2023neural}
Yuval Alaluf, Elad Richardson, Gal Metzer, and Daniel Cohen-Or.
\newblock A neural space-time representation for text-to-image personalization.
\newblock \emph{ACM Transactions on Graphics (TOG)}, 42\penalty0 (6):\penalty0 1--10, 2023.

\bibitem[Albergo and Vanden-Eijnden(2023)]{albergo2023building}
Michael~S. Albergo and Eric Vanden-Eijnden.
\newblock Building normalizing flows with stochastic interpolants.
\newblock In \emph{Proceedings of the International Conference on Learning Representations (ICLR) 2023}, 2023.
\newblock Also available as arXiv:2209.15571v3.

\bibitem[AlimamaCreative(2024)]{alimama2024control}
AlimamaCreative.
\newblock Flux-controlnet-inpainting. https://huggingface.co/alimama-creative/flux.1-dev-controlnet-inpainting-beta, 2024.
\newblock Accessed: 2025-01-30.

\bibitem[Avrahami et~al.(2022)Avrahami, Lischinski, and Fried]{avrahami2022blended}
Omri Avrahami, Dani Lischinski, and Ohad Fried.
\newblock Blended diffusion for text-driven editing of natural images.
\newblock In \emph{Proceedings of the IEEE/CVF conference on computer vision and pattern recognition}, pages 18208--18218, 2022.

\bibitem[Avrahami et~al.(2023)Avrahami, Fried, and Lischinski]{avrahami2023blended}
Omri Avrahami, Ohad Fried, and Dani Lischinski.
\newblock Blended latent diffusion.
\newblock \emph{ACM transactions on graphics (TOG)}, 42\penalty0 (4):\penalty0 1--11, 2023.

\bibitem[Baek()]{attention_map_diffusers}
Wooyeol Baek.
\newblock attention-map-diffusers.
\newblock \url{https://github.com/wooyeolbaek/attention-map-diffusers}.
\newblock Accessed: 07 Mar 2025.

\bibitem[Bertalmio et~al.(2000)Bertalmio, Sapiro, Caselles, and Ballester]{bertalmio2000inpainting}
Marcelo Bertalmio, Guillermo Sapiro, Vincent Caselles, and Coloma Ballester.
\newblock Image inpainting.
\newblock In \emph{Proceedings of the 27th Annual Conference on Computer Graphics and Interactive Techniques}, page 417–424, USA, 2000. ACM Press/Addison-Wesley Publishing Co.

\bibitem[Caron et~al.(2021)Caron, Touvron, Misra, J{\'e}gou, Mairal, Bojanowski, and Joulin]{caron2021emerging-dino}
Mathilde Caron, Hugo Touvron, Ishan Misra, Herv{\'e} J{\'e}gou, Julien Mairal, Piotr Bojanowski, and Armand Joulin.
\newblock Emerging properties in self-supervised vision transformers.
\newblock In \emph{Proceedings of the IEEE/CVF international conference on computer vision}, pages 9650--9660, 2021.

\bibitem[Chen et~al.(2022)Chen, Hu, Saharia, and Cohen]{chen2022re}
Wenhu Chen, Hexiang Hu, Chitwan Saharia, and William~W Cohen.
\newblock Re-imagen: Retrieval-augmented text-to-image generator.
\newblock \emph{arXiv preprint arXiv:2209.14491}, 2022.

\bibitem[Chen et~al.(2024{\natexlab{a}})Chen, Hu, Li, Ruiz, Jia, Chang, and Cohen]{chen2024subject}
Wenhu Chen, Hexiang Hu, Yandong Li, Nataniel Ruiz, Xuhui Jia, Ming-Wei Chang, and William~W Cohen.
\newblock Subject-driven text-to-image generation via apprenticeship learning.
\newblock \emph{Advances in Neural Information Processing Systems}, 36, 2024{\natexlab{a}}.

\bibitem[Chen et~al.(2024{\natexlab{b}})Chen, Huang, Liu, Shen, Zhao, and Zhao]{chen2024anydoor}
Xi Chen, Lianghua Huang, Yu Liu, Yujun Shen, Deli Zhao, and Hengshuang Zhao.
\newblock Anydoor: Zero-shot object-level image customization.
\newblock In \emph{Proceedings of the IEEE/CVF Conference on Computer Vision and Pattern Recognition}, pages 6593--6602, 2024{\natexlab{b}}.

\bibitem[Chen et~al.(2024{\natexlab{c}})Chen, Zhang, Zhang, Zhou, Kim, Liu, Li, Zhang, Zhao, Wang, et~al.]{chen2024unireal}
Xi Chen, Zhifei Zhang, He Zhang, Yuqian Zhou, Soo~Ye Kim, Qing Liu, Yijun Li, Jianming Zhang, Nanxuan Zhao, Yilin Wang, et~al.
\newblock Unireal: Universal image generation and editing via learning real-world dynamics.
\newblock \emph{arXiv preprint arXiv:2412.07774}, 2024{\natexlab{c}}.

\bibitem[Choi et~al.(2021)Choi, Park, Lee, and Choo]{choi2021viton}
Seunghwan Choi, Sunghyun Park, Minsoo Lee, and Jaegul Choo.
\newblock Viton-hd: High-resolution virtual try-on via misalignment-aware normalization.
\newblock In \emph{Proc. of the IEEE conference on computer vision and pattern recognition (CVPR)}, 2021.

\bibitem[Esser et~al.(2024)Esser, Kulal, Blattmann, Entezari, M{\"u}ller, Saini, Levi, Lorenz, Sauer, Boesel, et~al.]{esser2024scaling}
Patrick Esser, Sumith Kulal, Andreas Blattmann, Rahim Entezari, Jonas M{\"u}ller, Harry Saini, Yam Levi, Dominik Lorenz, Axel Sauer, Frederic Boesel, et~al.
\newblock Scaling rectified flow transformers for high-resolution image synthesis.
\newblock In \emph{Forty-first International Conference on Machine Learning}, 2024.

\bibitem[Fei et~al.(2023)Fei, Fan, and Huang]{fei2023gradient}
Zhengcong Fei, Mingyuan Fan, and Junshi Huang.
\newblock Gradient-free textual inversion.
\newblock In \emph{Proceedings of the 31st ACM International Conference on Multimedia}, pages 1364--1373, 2023.

\bibitem[Gal et~al.(2022)Gal, Alaluf, Atzmon, Patashnik, Bermano, Chechik, and Cohen-Or]{gal2022image}
Rinon Gal, Yuval Alaluf, Yuval Atzmon, Or Patashnik, Amit~H Bermano, Gal Chechik, and Daniel Cohen-Or.
\newblock An image is worth one word: Personalizing text-to-image generation using textual inversion.
\newblock \emph{arXiv preprint arXiv:2208.01618}, 2022.

\bibitem[Goodfellow et~al.(2014)Goodfellow, Pouget-Abadie, Mirza, Xu, Warde-Farley, Ozair, Courville, and Bengio]{goodfellow2014generative}
Ian Goodfellow, Jean Pouget-Abadie, Mehdi Mirza, Bing Xu, David Warde-Farley, Sherjil Ozair, Aaron Courville, and Yoshua Bengio.
\newblock Generative adversarial nets.
\newblock \emph{Advances in neural information processing systems}, 27, 2014.

\bibitem[Hao et~al.(2023)Hao, Chi, Dong, and Wei]{hao2023optimizingpromptstexttoimagegeneration}
Yaru Hao, Zewen Chi, Li Dong, and Furu Wei.
\newblock Optimizing prompts for text-to-image generation, 2023.

\bibitem[He et~al.(2019)He, Zuo, Kan, Shan, and Chen]{he2019attgan}
Zhenliang He, Wangmeng Zuo, Meina Kan, Shiguang Shan, and Xilin Chen.
\newblock Attgan: Facial attribute editing by only changing what you want.
\newblock \emph{IEEE transactions on image processing}, 28\penalty0 (11):\penalty0 5464--5478, 2019.

\bibitem[Ho et~al.(2020)Ho, Jain, and Abbeel]{ho2020denoising}
Jonathan Ho, Ajay Jain, and Pieter Abbeel.
\newblock Denoising diffusion probabilistic models.
\newblock \emph{Advances in neural information processing systems}, 33:\penalty0 6840--6851, 2020.

\bibitem[Hu et~al.(2021)Hu, Shen, Wallis, Allen-Zhu, Li, Wang, Wang, and Chen]{hu2021lora}
Edward~J Hu, Yelong Shen, Phillip Wallis, Zeyuan Allen-Zhu, Yuanzhi Li, Shean Wang, Lu Wang, and Weizhu Chen.
\newblock Lora: Low-rank adaptation of large language models.
\newblock \emph{arXiv preprint arXiv:2106.09685}, 2021.

\bibitem[Huang et~al.(2024)Huang, Wang, Wu, Shi, Dou, Liang, Feng, Liu, and Zhou]{lhhuang2024iclora}
Lianghua Huang, Wei Wang, Zhi-Fan Wu, Yupeng Shi, Huanzhang Dou, Chen Liang, Yutong Feng, Yu Liu, and Jingren Zhou.
\newblock In-context lora for diffusion transformers.
\newblock \emph{arXiv preprint arxiv:2410.23775}, 2024.

\bibitem[Hui et~al.(2024)Hui, Yang, Zhao, Shi, Wang, Wang, Zhou, and Xie]{hui2024hq}
Mude Hui, Siwei Yang, Bingchen Zhao, Yichun Shi, Heng Wang, Peng Wang, Yuyin Zhou, and Cihang Xie.
\newblock Hq-edit: A high-quality dataset for instruction-based image editing.
\newblock \emph{arXiv preprint arXiv:2404.09990}, 2024.

\bibitem[Jia et~al.(2023)Jia, Zhao, Chan, Li, Zhang, Gong, Hou, Wang, and Su]{jia2023taming}
Xuhui Jia, Yang Zhao, Kelvin~CK Chan, Yandong Li, Han Zhang, Boqing Gong, Tingbo Hou, Huisheng Wang, and Yu-Chuan Su.
\newblock Taming encoder for zero fine-tuning image customization with text-to-image diffusion models.
\newblock \emph{arXiv preprint arXiv:2304.02642}, 2023.

\bibitem[Ju et~al.(2024)Ju, Liu, Wang, Bian, Shan, and Xu]{ju2024brushnet}
Xuan Ju, Xian Liu, Xintao Wang, Yuxuan Bian, Ying Shan, and Qiang Xu.
\newblock Brushnet: A plug-and-play image inpainting model with decomposed dual-branch diffusion.
\newblock In \emph{European Conference on Computer Vision}, pages 150--168. Springer, 2024.

\bibitem[Karras et~al.(2021)Karras, Laine, and Aila]{tero2021style}
Tero Karras, Samuli Laine, and Timo Aila.
\newblock A style-based generator architecture for generative adversarial networks.
\newblock \emph{IEEE Transactions on Pattern Analysis and Machine Intelligence}, 43\penalty0 (12):\penalty0 4217--4228, 2021.

\bibitem[Kumari et~al.(2023)Kumari, Zhang, Zhang, Shechtman, and Zhu]{kumari2023multi}
Nupur Kumari, Bingliang Zhang, Richard Zhang, Eli Shechtman, and Jun-Yan Zhu.
\newblock Multi-concept customization of text-to-image diffusion.
\newblock In \emph{Proceedings of the IEEE/CVF Conference on Computer Vision and Pattern Recognition}, pages 1931--1941, 2023.

\bibitem[Labs(2024{\natexlab{a}})]{black2024fill}
Black~Forest Labs.
\newblock Flux.1-fill-dev.
\newblock https://huggingface.co/black-forest-labs/FLUX.1-Fill-dev, 2024{\natexlab{a}}.
\newblock Accessed: 2025-01-30.

\bibitem[Labs(2024{\natexlab{b}})]{black2024flux}
Black~Forest Labs.
\newblock Flux.1-dev.
\newblock https://huggingface.co/black-forest-labs/FLUX.1-dev, 2024{\natexlab{b}}.
\newblock Accessed: 2025-01-30.

\bibitem[Li et~al.(2024)Li, Li, and Hoi]{li2024blip}
Dongxu Li, Junnan Li, and Steven Hoi.
\newblock Blip-diffusion: Pre-trained subject representation for controllable text-to-image generation and editing.
\newblock \emph{Advances in Neural Information Processing Systems}, 36, 2024.

\bibitem[Lipman et~al.(2023)Lipman, Chen, Ben-Hamu, Nickel, and Le]{lipman2023flowmatching}
Yaron Lipman, Ricky T.~Q. Chen, Heli Ben-Hamu, Maximilian Nickel, and Matt Le.
\newblock Flow matching for generative modeling.
\newblock In \emph{Proceedings of the International Conference on Learning Representations (ICLR) 2023}, 2023.
\newblock Also available as arXiv:2210.02747v2.

\bibitem[Liu et~al.(2021)Liu, Wan, Huang, Song, Han, and Liao]{liu2021pd}
Hongyu Liu, Ziyu Wan, Wei Huang, Yibing Song, Xintong Han, and Jing Liao.
\newblock Pd-gan: Probabilistic diverse gan for image inpainting.
\newblock In \emph{Proceedings of the IEEE/CVF conference on computer vision and pattern recognition}, pages 9371--9381, 2021.

\bibitem[Liu et~al.(2023{\natexlab{a}})Liu, Gong, and Liu]{liu2023flowstraight}
Xingchao Liu, Chengyue Gong, and Qiang Liu.
\newblock Flow straight and fast: Learning to generate and transfer data with rectified flow.
\newblock In \emph{Proceedings of the International Conference on Learning Representations (ICLR) 2023}, 2023{\natexlab{a}}.
\newblock Also available as arXiv:2209.03003.

\bibitem[Liu et~al.(2023{\natexlab{b}})Liu, Feng, Zhu, Zhang, Zheng, Liu, Zhao, Zhou, and Cao]{liu2023cones}
Zhiheng Liu, Ruili Feng, Kai Zhu, Yifei Zhang, Kecheng Zheng, Yu Liu, Deli Zhao, Jingren Zhou, and Yang Cao.
\newblock Cones: Concept neurons in diffusion models for customized generation.
\newblock \emph{arXiv preprint arXiv:2303.05125}, 2023{\natexlab{b}}.

\bibitem[Lugmayr et~al.(2022)Lugmayr, Danelljan, Romero, Yu, Timofte, and Van~Gool]{lugmayr2022repaint}
Andreas Lugmayr, Martin Danelljan, Andres Romero, Fisher Yu, Radu Timofte, and Luc Van~Gool.
\newblock Repaint: Inpainting using denoising diffusion probabilistic models.
\newblock In \emph{Proceedings of the IEEE/CVF Conference on Computer Vision and Pattern Recognition (CVPR)}, pages 11461--11471, 2022.

\bibitem[Ma et~al.(2024)Ma, Liang, Chen, and Lu]{ma2024subject}
Jian Ma, Junhao Liang, Chen Chen, and Haonan Lu.
\newblock Subject-diffusion: Open domain personalized text-to-image generation without test-time fine-tuning.
\newblock In \emph{ACM SIGGRAPH 2024 Conference Papers}, pages 1--12, 2024.

\bibitem[OpenAI(2023)]{openai2023dalle3}
OpenAI.
\newblock Dall·e 3.
\newblock \url{https://openai.com/dall-e-3}, 2023.
\newblock Accessed: 2025-01-30.

\bibitem[Pan et~al.(2023)Pan, Dong, Huang, Peng, Chen, and Wei]{pan2023kosmos}
Xichen Pan, Li Dong, Shaohan Huang, Zhiliang Peng, Wenhu Chen, and Furu Wei.
\newblock Kosmos-g: Generating images in context with multimodal large language models.
\newblock \emph{arXiv preprint arXiv:2310.02992}, 2023.

\bibitem[Patel et~al.(2024)Patel, Jung, Baral, and Yang]{patel2024lambda}
Maitreya Patel, Sangmin Jung, Chitta Baral, and Yezhou Yang.
\newblock $\lambda$-eclipse: Multi-concept personalized text-to-image diffusion models by leveraging clip latent space.
\newblock \emph{arXiv preprint arXiv:2402.05195}, 2024.

\bibitem[Peng et~al.(2021)Peng, Liu, Xu, and Li]{peng2021generating}
Jialun Peng, Dong Liu, Songcen Xu, and Houqiang Li.
\newblock Generating diverse structure for image inpainting with hierarchical vq-vae.
\newblock In \emph{Proceedings of the IEEE/CVF conference on computer vision and pattern recognition}, pages 10775--10784, 2021.

\bibitem[Podell et~al.(2023)Podell, English, Lacey, Blattmann, Dockhorn, M{\"u}ller, Penna, and Rombach]{podell2023sdxl}
Dustin Podell, Zion English, Kyle Lacey, Andreas Blattmann, Tim Dockhorn, Jonas M{\"u}ller, Joe Penna, and Robin Rombach.
\newblock Sdxl: Improving latent diffusion models for high-resolution image synthesis.
\newblock \emph{arXiv preprint arXiv:2307.01952}, 2023.

\bibitem[Quan et~al.(2024)Quan, Chen, Liu, Yan, and Wonka]{quan2024deep}
Weize Quan, Jiaxi Chen, Yanli Liu, Dong-Ming Yan, and Peter Wonka.
\newblock Deep learning-based image and video inpainting: A survey.
\newblock \emph{International Journal of Computer Vision}, 132\penalty0 (7):\penalty0 2367--2400, 2024.

\bibitem[Radford et~al.(2021)Radford, Kim, Hallacy, Ramesh, Goh, Agarwal, Sastry, Askell, Mishkin, Clark, et~al.]{radford2021learning-clip}
Alec Radford, Jong~Wook Kim, Chris Hallacy, Aditya Ramesh, Gabriel Goh, Sandhini Agarwal, Girish Sastry, Amanda Askell, Pamela Mishkin, Jack Clark, et~al.
\newblock Learning transferable visual models from natural language supervision.
\newblock In \emph{International conference on machine learning}, pages 8748--8763. PMLR, 2021.

\bibitem[Richardson et~al.(2021)Richardson, Alaluf, Patashnik, Nitzan, Azar, Shapiro, and Cohen-Or]{richardson2021encoding}
Elad Richardson, Yuval Alaluf, Or Patashnik, Yotam Nitzan, Yaniv Azar, Stav Shapiro, and Daniel Cohen-Or.
\newblock Encoding in style: a stylegan encoder for image-to-image translation.
\newblock In \emph{Proceedings of the IEEE/CVF conference on computer vision and pattern recognition}, pages 2287--2296, 2021.

\bibitem[Rombach et~al.(2022)Rombach, Blattmann, Lorenz, Esser, and Ommer]{rombach2022high}
Robin Rombach, Andreas Blattmann, Dominik Lorenz, Patrick Esser, and Bj{\"o}rn Ommer.
\newblock High-resolution image synthesis with latent diffusion models.
\newblock In \emph{Proceedings of the IEEE/CVF conference on computer vision and pattern recognition}, pages 10684--10695, 2022.

\bibitem[Ronneberger et~al.(2015)Ronneberger, Fischer, and Brox]{ronneberger2015u}
Olaf Ronneberger, Philipp Fischer, and Thomas Brox.
\newblock U-net: Convolutional networks for biomedical image segmentation.
\newblock In \emph{Medical image computing and computer-assisted intervention--MICCAI 2015: 18th international conference, Munich, Germany, October 5-9, 2015, proceedings, part III 18}, pages 234--241. Springer, 2015.

\bibitem[Rout et~al.(2024)Rout, Chen, Ruiz, Caramanis, Shakkottai, and Chu]{rout2024semanticimageinversionediting}
Litu Rout, Yujia Chen, Nataniel Ruiz, Constantine Caramanis, Sanjay Shakkottai, and Wen-Sheng Chu.
\newblock Semantic image inversion and editing using rectified stochastic differential equations, 2024.

\bibitem[Ruiz et~al.(2023)Ruiz, Li, Jampani, Pritch, Rubinstein, and Aberman]{ruiz2023dreambooth}
Nataniel Ruiz, Yuanzhen Li, Varun Jampani, Yael Pritch, Michael Rubinstein, and Kfir Aberman.
\newblock Dreambooth: Fine tuning text-to-image diffusion models for subject-driven generation.
\newblock In \emph{Proceedings of the IEEE/CVF conference on computer vision and pattern recognition}, pages 22500--22510, 2023.

\bibitem[Shi et~al.(2024)Shi, Xiong, Lin, and Jung]{shi2024instantbooth}
Jing Shi, Wei Xiong, Zhe Lin, and Hyun~Joon Jung.
\newblock Instantbooth: Personalized text-to-image generation without test-time finetuning.
\newblock In \emph{Proceedings of the IEEE/CVF Conference on Computer Vision and Pattern Recognition}, pages 8543--8552, 2024.

\bibitem[Shin et~al.(2024)Shin, Choi, Kim, and Yoon]{shin2024diptych}
Chaehun Shin, Jooyoung Choi, Heeseung Kim, and Sungroh Yoon.
\newblock Large-scale text-to-image model with inpainting is a zero-shot subject-driven image generator.
\newblock \emph{arXiv preprint arxiv:2411.15466}, 2024.

\bibitem[Shoshan et~al.(2021)Shoshan, Bhonker, Kviatkovsky, and Medioni]{shoshan2021gancontrol}
Alon Shoshan, Nadav Bhonker, Igor Kviatkovsky, and G\'erard Medioni.
\newblock Gan-control: Explicitly controllable gans.
\newblock In \emph{Proceedings of the IEEE/CVF International Conference on Computer Vision (ICCV)}, pages 14083--14093, 2021.

\bibitem[Song et~al.(2020)Song, Meng, and Ermon]{song2020denoising}
Jiaming Song, Chenlin Meng, and Stefano Ermon.
\newblock Denoising diffusion implicit models.
\newblock \emph{arXiv preprint arXiv:2010.02502}, 2020.

\bibitem[Song et~al.(2024)Song, Zhang, Lin, Cohen, Price, Zhang, Kim, Zhang, Xiong, and Aliaga]{song2024imprint}
Yizhi Song, Zhifei Zhang, Zhe Lin, Scott Cohen, Brian Price, Jianming Zhang, Soo~Ye Kim, He Zhang, Wei Xiong, and Daniel Aliaga.
\newblock Imprint: Generative object compositing by learning identity-preserving representation.
\newblock In \emph{Proceedings of the IEEE/CVF Conference on Computer Vision and Pattern Recognition}, pages 8048--8058, 2024.

\bibitem[Sun et~al.(2024)Sun, Cui, Zhang, Zhang, Yu, Wang, Rao, Liu, Huang, and Wang]{sun2024generative}
Quan Sun, Yufeng Cui, Xiaosong Zhang, Fan Zhang, Qiying Yu, Yueze Wang, Yongming Rao, Jingjing Liu, Tiejun Huang, and Xinlong Wang.
\newblock Generative multimodal models are in-context learners.
\newblock In \emph{Proceedings of the IEEE/CVF Conference on Computer Vision and Pattern Recognition}, pages 14398--14409, 2024.

\bibitem[Sutskever(2024)]{ilyas2024pretraining}
Ilya Sutskever.
\newblock Sequence to sequence learning with neural networks: what a decade.
\newblock Conference talk at Neural Information Processing Systems (NeurIPS) 2024, 2024.

\bibitem[Tan et~al.(2024)Tan, Liu, Yang, Xue, and Wang]{tan2024ominicontrol}
Zhenxiong Tan, Songhua Liu, Xingyi Yang, Qiaochu Xue, and Xinchao Wang.
\newblock Ominicontrol: Minimal and universal control for diffusion transformer.
\newblock \emph{arXiv preprint arXiv:2411.15098}, 3, 2024.

\bibitem[Voynov et~al.(2023)Voynov, Chu, Cohen-Or, and Aberman]{voynov2023p+}
Andrey Voynov, Qinghao Chu, Daniel Cohen-Or, and Kfir Aberman.
\newblock p+: Extended textual conditioning in text-to-image generation.
\newblock \emph{arXiv preprint arXiv:2303.09522}, 2023.

\bibitem[Wang et~al.(2024{\natexlab{a}})Wang, Fu, Huang, He, and Jiang]{wang2024ms}
X Wang, Siming Fu, Qihan Huang, Wanggui He, and Hao Jiang.
\newblock Ms-diffusion: Multi-subject zero-shot image personalization with layout guidance.
\newblock \emph{arXiv preprint arXiv:2406.07209}, 2024{\natexlab{a}}.

\bibitem[Wang et~al.(2024{\natexlab{b}})Wang, Huang, Song, Ma, and Zhang]{Wang_2024}
Zhijie Wang, Yuheng Huang, Da Song, Lei Ma, and Tianyi Zhang.
\newblock Promptcharm: Text-to-image generation through multi-modal prompting and refinement.
\newblock In \emph{Proceedings of the CHI Conference on Human Factors in Computing Systems}, page 1–21. ACM, 2024{\natexlab{b}}.

\bibitem[Wei et~al.(2023)Wei, Zhang, Ji, Bai, Zhang, and Zuo]{wei2023elite}
Yuxiang Wei, Yabo Zhang, Zhilong Ji, Jinfeng Bai, Lei Zhang, and Wangmeng Zuo.
\newblock Elite: Encoding visual concepts into textual embeddings for customized text-to-image generation.
\newblock In \emph{Proceedings of the IEEE/CVF International Conference on Computer Vision}, pages 15943--15953, 2023.

\bibitem[Winter et~al.(2024)Winter, Shul, Cohen, Berman, Pritch, Rav-Acha, and Hoshen]{winter2024objectmate}
Daniel Winter, Asaf Shul, Matan Cohen, Dana Berman, Yael Pritch, Alex Rav-Acha, and Yedid Hoshen.
\newblock Objectmate: A recurrence prior for object insertion and subject-driven generation.
\newblock \emph{arXiv preprint arXiv:2412.08645}, 2024.

\bibitem[Xie et~al.(2023)Xie, Zhang, Lin, Hinz, and Zhang]{xie2023smartbrush}
Shaoan Xie, Zhifei Zhang, Zhe Lin, Tobias Hinz, and Kun Zhang.
\newblock Smartbrush: Text and shape guided object inpainting with diffusion model.
\newblock In \emph{Proceedings of the IEEE/CVF Conference on Computer Vision and Pattern Recognition}, pages 22428--22437, 2023.

\bibitem[Ye et~al.(2023)Ye, Zhang, Liu, Han, and Yang]{ye2023ip}
Hu Ye, Jun Zhang, Sibo Liu, Xiao Han, and Wei Yang.
\newblock Ip-adapter: Text compatible image prompt adapter for text-to-image diffusion models.
\newblock \emph{arXiv preprint arXiv:2308.06721}, 2023.

\bibitem[Zeng et~al.(2024)Zeng, Patel, Wang, Huang, Wang, Liu, and Balaji]{zeng2024jedi}
Yu Zeng, Vishal~M Patel, Haochen Wang, Xun Huang, Ting-Chun Wang, Ming-Yu Liu, and Yogesh Balaji.
\newblock Jedi: Joint-image diffusion models for finetuning-free personalized text-to-image generation.
\newblock In \emph{Proceedings of the IEEE/CVF Conference on Computer Vision and Pattern Recognition}, pages 6786--6795, 2024.

\bibitem[Zhang et~al.(2023)Zhang, Rao, and Agrawala]{zhang2023adding}
Lvmin Zhang, Anyi Rao, and Maneesh Agrawala.
\newblock Adding conditional control to text-to-image diffusion models, 2023.

\bibitem[Zhang et~al.(2024)Zhang, Song, Liu, Wang, Yu, Tang, Li, Tang, Hu, Pan, et~al.]{zhang2024ssr}
Yuxuan Zhang, Yiren Song, Jiaming Liu, Rui Wang, Jinpeng Yu, Hao Tang, Huaxia Li, Xu Tang, Yao Hu, Han Pan, et~al.
\newblock Ssr-encoder: Encoding selective subject representation for subject-driven generation.
\newblock In \emph{Proceedings of the IEEE/CVF Conference on Computer Vision and Pattern Recognition}, pages 8069--8078, 2024.

\bibitem[Zhao et~al.(2017)Zhao, Xiong, Karlekar~Jayashree, Li, Zhao, Wang, Sugiri~Pranata, Shengmei~Shen, Yan, and Feng]{jian2017dual}
Jian Zhao, Lin Xiong, Panasonic Karlekar~Jayashree, Jianshu Li, Fang Zhao, Zhecan Wang, Panasonic Sugiri~Pranata, Panasonic Shengmei~Shen, Shuicheng Yan, and Jiashi Feng.
\newblock Dual-agent gans for photorealistic and identity preserving profile face synthesis.
\newblock In \emph{Advances in Neural Information Processing Systems}. Curran Associates, Inc., 2017.

\bibitem[Zheng et~al.(2019)Zheng, Cham, and Cai]{zheng2019pluralistic}
Chuanxia Zheng, Tat-Jen Cham, and Jianfei Cai.
\newblock Pluralistic image completion.
\newblock In \emph{Proceedings of the IEEE/CVF Conference on Computer Vision and Pattern Recognition}, pages 1438--1447, 2019.

\bibitem[Zheng et~al.(2022)Zheng, Lin, Lu, Cohen, Shechtman, Barnes, Zhang, Xu, Amirghodsi, and Luo]{zheng2022image}
Haitian Zheng, Zhe Lin, Jingwan Lu, Scott Cohen, Eli Shechtman, Connelly Barnes, Jianming Zhang, Ning Xu, Sohrab Amirghodsi, and Jiebo Luo.
\newblock Image inpainting with cascaded modulation gan and object-aware training.
\newblock In \emph{European Conference on Computer Vision}, pages 277--296. Springer, 2022.

\bibitem[Zheng et~al.(2024)Zheng, Gao, Fan, Liu, Laaksonen, Ouyang, and Sebe]{BiRefNet}
Peng Zheng, Dehong Gao, Deng-Ping Fan, Li Liu, Jorma Laaksonen, Wanli Ouyang, and Nicu Sebe.
\newblock Bilateral reference for high-resolution dichotomous image segmentation.
\newblock \emph{CAAI Artificial Intelligence Research}, 2024.

\bibitem[Zhu et~al.(2024)Zhu, Chen, Wang, Zhao, and Jia]{zhu2024logosticker}
Mingkang Zhu, Xi Chen, Zhongdao Wang, Hengshuang Zhao, and Jiaya Jia.
\newblock Logosticker: Inserting logos into diffusion models for customized generation.
\newblock In \emph{European Conference on Computer Vision}, pages 363--378. Springer, 2024.

\end{thebibliography}

\clearpage

\section{Appendix}
\subsection{Background}

\paragraph{Diffusion Models}
\label{sec:app:diffusion_models}

Diffusion models \cite{ho2020denoising, song2020denoising} define a generative process that gradually transforms data from a complex distribution into a simple noise distribution and then learns to reverse this process. The \emph{forward process} is typically formulated as a stochastic differential equation (SDE):
\begin{equation} \label{eq:forward_sde}
  dx_t = f(x_t, t) \, dt + g(t) \, dW_t, \quad t \in [0, T],
\end{equation}
where:
\begin{itemize}
  \item $x_t \in \mathbb{R}^d$ is the state at time $t$,
  \item $f(x_t, t)$ is a drift term,
  \item $g(t)$ is a time-dependent diffusion coefficient,
  \item $W_t$ denotes a standard Wiener process.
\end{itemize}
At $t=0$, the data sample is drawn from the target distribution, \( x_0 \sim p_{\text{data}}(x) \), and as \( t \) increases, noise is added until \( x_T \) approximates a simple distribution (e.g., a Gaussian).

The \emph{reverse process} aims to recover the data by inverting the noising dynamics. Its dynamics are described by the reverse-time SDE:
\begin{equation} \label{eq:reverse_sde}
  dx_t = \left[f(x_t, t) - g(t)^2 \nabla_{x_t} \log p_t(x_t)\right] dt + g(t) \, d\overline{W}_t,
\end{equation}
where:
\begin{itemize}
  \item \( p_t(x) \) is the marginal density at time \( t \),
  \item \( \nabla_{x_t} \log p_t(x_t) \) is the score function,
  \item \( d\overline{W}_t \) is a reverse-time Wiener process.
\end{itemize}
Training typically focuses on estimating the score function \( \nabla_{x_t} \log p_t(x_t) \) via techniques such as denoising score matching \cite{song2020denoising}, other equivalent parametrization like the denoiser have also been proposed \cite{ho2020denoising}. Once a good approximation is obtained, samples can be generated by simulating the reverse SDE from \( t = T \) back to \( t = 0 \).

\paragraph{Flow Matching}
\label{sec:app:flow-matching}
Flow matching is an alternative approach that seeks to learn a deterministic flow to transport a simple base distribution $p_0(x)$ into the target distribution $p_T(x)$ \cite{lipman2023flowmatching, liu2023flowstraight, albergo2023building}. Instead of reversing a stochastic process, one directly learns a time-dependent vector field $\mathbf{v}(x,t)$ that governs this transformation.

The evolution of the probability density $p_t(x)$ under a deterministic flow is governed by the continuity equation:
$$
\frac{\partial p_t(x)}{\partial t} + \nabla \cdot \Bigl(p_t(x) \, \mathbf{v}(x,t)\Bigr) = 0,
$$
with the boundary conditions $p_0(x)$ at $t=0$ and $p_T(x) \approx p_{\text{data}}(x)$ at $t=T$. The deterministic dynamics that transport samples along this flow are given by the ordinary differential equation (ODE):
$$
\frac{d x_t}{dt} = \mathbf{v}(x_t, t).
$$
Starting with $x_0 \sim p_0(x)$ and integrating this ODE until $t=T$, the resulting $x_T$ should follow the target distribution.

In flow matching, a parameterized vector field $\mathbf{v}_\theta(x,t)$ is learned by minimizing a loss that measures its discrepancy from an ideal (or reference) flow $\mathbf{v}^*(x,t)$. A common choice is the squared-error loss:
$$
\mathcal{L}(\theta) = \mathbb{E}_{t,x_t} \left[ \left\| \mathbf{v}_\theta(x_t,t) - \mathbf{v}^*(x_t,t) \right\|^2 \right].
$$

The reference flow $\mathbf{v}^*(x,t)$ is often derived either from the dynamics of a corresponding forward process or directly from the optimal coupling between the base and target distributions. In practice, since the base distribution is typically chosen as a Gaussian, its analytical properties can be exploited to obtain a closed-form expression.

\subsection{More on Mosaic Attention Visualizations}
\begin{figure*}[ht]
    \centering
    \includegraphics[width=\textwidth, trim=0 270 0 0, clip]{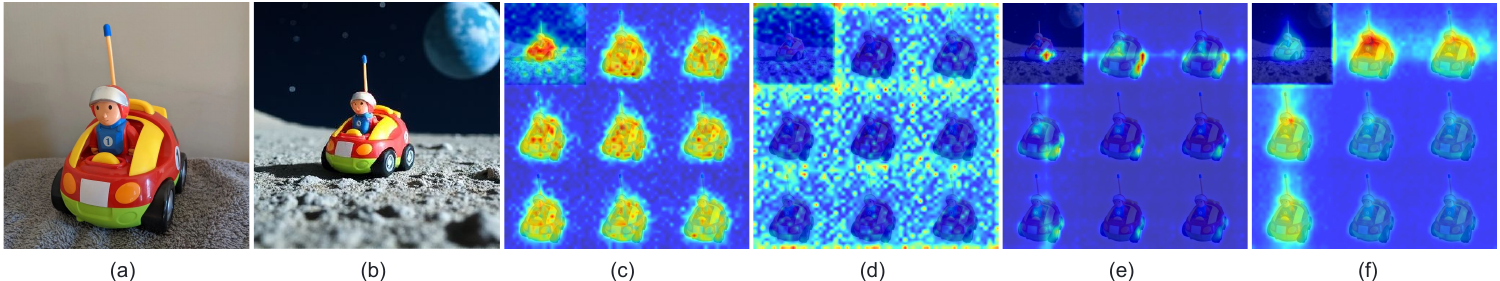}
    \caption{Attention visualization with an RC car on mosaic pattern. (a) The reference image featuring the subject; (b) The generated subject image; (c) The attention visualization for the text tokens "a colorful toy car with a driver figure"; (d) The attention visualization for the text tokens "this set of images"; (e) The attention visualization for a single token on the frontal left wheel; (f) The attention visualization for the subject-related tokens.}
    \label{fig:att_viz}
\end{figure*}

More attention visualizations of the mosaic pattern are shown in~\ref{fig:att_viz}, using an RC car subject with the text prompt: \textit{"this set of images presents a colorful toy car with a driver figure. [IMAGE1] features a film-style shot. On the moon, the item drives across the moon's surface. The background shows Earth looming large in the foreground. [IMAGE2] highlights the item's vibrant colors and playful design..."}

\subsection{More on Human Preference Study}
\begin{figure*}[ht]
    \centering
    \includegraphics[width=0.7\textwidth, trim=0 0 350 0, clip]{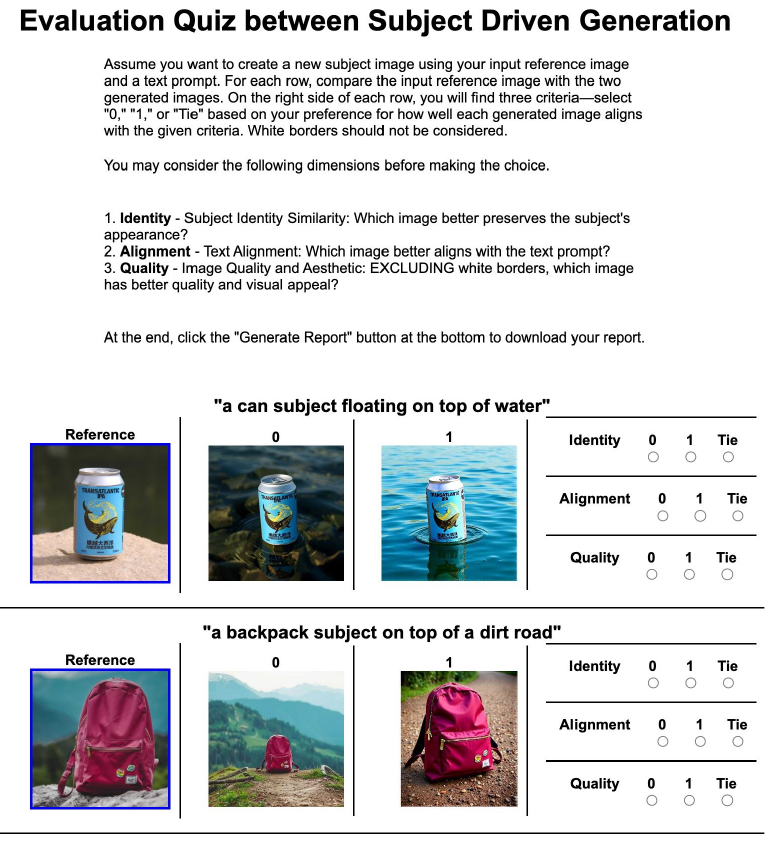}
    \caption{A screenshot of our Human Preference Study questionnaire.}
    \label{fig:user_study}
\end{figure*}

We present a screenshot of our Human Preference Study questionnaire in~\ref{fig:user_study}.

\subsection{More on Quantitative Comparisons}
Table~\ref{tab:sup:baseline_comparison_main} presents additional comparisons with baseline methods.
\begin{table*}[t]
\centering
\caption{Comparison of models with their respective metrics. In all metrics higher is better. SD- and SDXL-based methods generally achieve higher CLIP scores due to their direct integration of CLIP within the architecture. $^\spadesuit$ denotes DreamBooth was tuned on top of the original method. Results copied from the original papers except for $^\clubsuit$ denotes that results come from the evaluation of the method in \cite{patel2024lambda} while $^\diamondsuit$ was evaluated in \cite{shin2024diptych}, 
$^a$ While Re-Image doesn't require any training, it relies on a database of images from which it retrieves reference images. $^\heartsuit$ is re-implemented in this work. The scores are shown in percentage.}
\label{tab:sup:baseline_comparison_main}
\resizebox{0.99\textwidth}{!}{%
\begin{tabular}{clcccccc}
\toprule
\textbf{Type} & \textbf{Method} & \textbf{Base} & \textbf{Params} & \textbf{GPU-H} & \textbf{CLIP-I$\uparrow$} & \textbf{DINO$\uparrow$} & \textbf{CLIP-T$\uparrow$} \\
\midrule
\multirow{5}{*}{\shortstack{Test-time\\tuning\\w/\\extra params}}
& Text. Inversion \cite{gal2022image}   & SD 1.5 & 768  & 1    & 78.0 & 56.9 & 25.5 \\
& DreamBooth \cite{ruiz2023dreambooth}        & SD 1.5 & 0.9B & 0.2  & 80.3 & 66.8 & 30.5 \\
& BLIP-Diff~$^\spadesuit$  \cite{li2024blip}        & SD 1.5 & 0.9B & 0.1  & 80.5 & 67.0 & 30.2 \\
& Custom-Diff \cite{kumari2023multi}   & SD 1.5 & 57M  & 0.2  & 79.0 & 64.3 & 30.5 \\
& $\lambda$-ECLIPSE~$^\spadesuit$ \cite{patel2024lambda} & Kan 2.2  & 0.9B & 0.2  & 79.6 & 68.2 & 30.4 \\
\midrule
\multirow{17}{*}{\shortstack{Pre-trained \\ w/\\extra params}}
& ELITE \cite{wei2023elite}              & SD 1.4 & 77M  & 336  & 77.1 & 62.1 & 29.3 \\
& Subject-Diff \cite{ma2024subject}      & SD 1.5 & 252M & $n/a$   & 78.7 & 71.1 & 29.3 \\
& BLIP-Diff \cite{li2024blip}          & SD 1.5 & 1.5B & 2304 & 77.9 & 59.4 & 30.0 \\
& BLIP-Diff $^\clubsuit$ \cite{li2024blip}         & SD 1.5 & 1.5B & 2304 & 79.3 & 60.3 & 29.1 \\
\cdashline{2-8}[1pt/2pt]
\noalign{\vskip 3pt}   
& IP-Adapter $^\clubsuit$ \cite{ye2023ip}      & SD 1.5 & 22M  & 672  & 82.7 & 62.9 & 26.4 \\
& IP-Adapter $^\clubsuit$ \cite{ye2023ip}       & SDXL   & 22M  & 672  & 81.0 & 61.3 & 29.2 \\
& IP-Adapter $^\diamondsuit$ \cite{ye2023ip}      & FLUX & 22M  & 672 & 72.5 & 56.1 & 35.1 \\
\cdashline{2-8}[1pt/2pt]
\noalign{\vskip 3pt}   
& Kosmos-G  \cite{pan2023kosmos}  & SD 1.5 & 1.9B & 12300   & 84.7 & 69.4 & 28.7 \\
& Kosmos-G $^\clubsuit$ \cite{pan2023kosmos}         & SD 1.5 & 1.9B & 12300 & 82.2 & 61.8 & 25.0 \\
& MS-Diff \cite{wang2024ms} & SDXL & $n/a$ & $n/a$ & 79.2 & 67.1 & 32.1 \\
& Emu2 \cite{sun2024generative}  &SDXL & 37B & $n/a$ &  85.0 & 76.6  & 28.7 \\
& Emu2 $^\clubsuit$ \cite{sun2024generative}              & SDXL   & 37B  & $n/a$    & 76.5 & 56.3 & 27.3 \\
& $\lambda$-ECLIPSE \cite{patel2024lambda} & Kan 2.2  & 34M  & 74    & 78.3 & 61.3 & 30.7 \\
& OminiControl \cite{tan2024ominicontrol}              & FLUX   & 14.5M  & $n/a$    & 77.3 & 62.7 & 32.2 \\
& Diptych \cite{shin2024diptych} & FLUX & 2B & $n/a$ & 75.8 & 68.8 & 34.5 \\
& Diptych$^{\heartsuit}$ \cite{shin2024diptych} & FLUX & 2B & $n/a$ & 79.4 & 66.6 & 31.9 \\
\midrule
\multirow{4}{*}{Param-free}
& Re-Imagen \cite{chen2022re}          & Imagen & \multicolumn{2}{c}{see $^a$}   & 74.0 & 60.0 & 27.0 \\
& RF-Inversion \cite{rout2024semanticimageinversionediting} & FLUX & -- & -- & 78.7 & 61.9 & 29.4 \\
& LatentUnfold (1 view) & FLUX & -- & --   & 78.7 & 64.0 & 30.5 \\
& LatentUnfold (3 views\textsuperscript{avg})\ & FLUX & -- & --   &  77.6 & 61.8 & 30.5 \\
& LatentUnfold (3 views\textsuperscript{best}) & FLUX & -- & --   &  80.6 & 66.0 & 30.5 \\
\bottomrule
\end{tabular}}
\end{table*}

\subsection{More on Qualitative Results}
\begin{figure*}[ht]
    \centering
    \includegraphics[width=\textwidth, trim=0 0 0 0, clip]{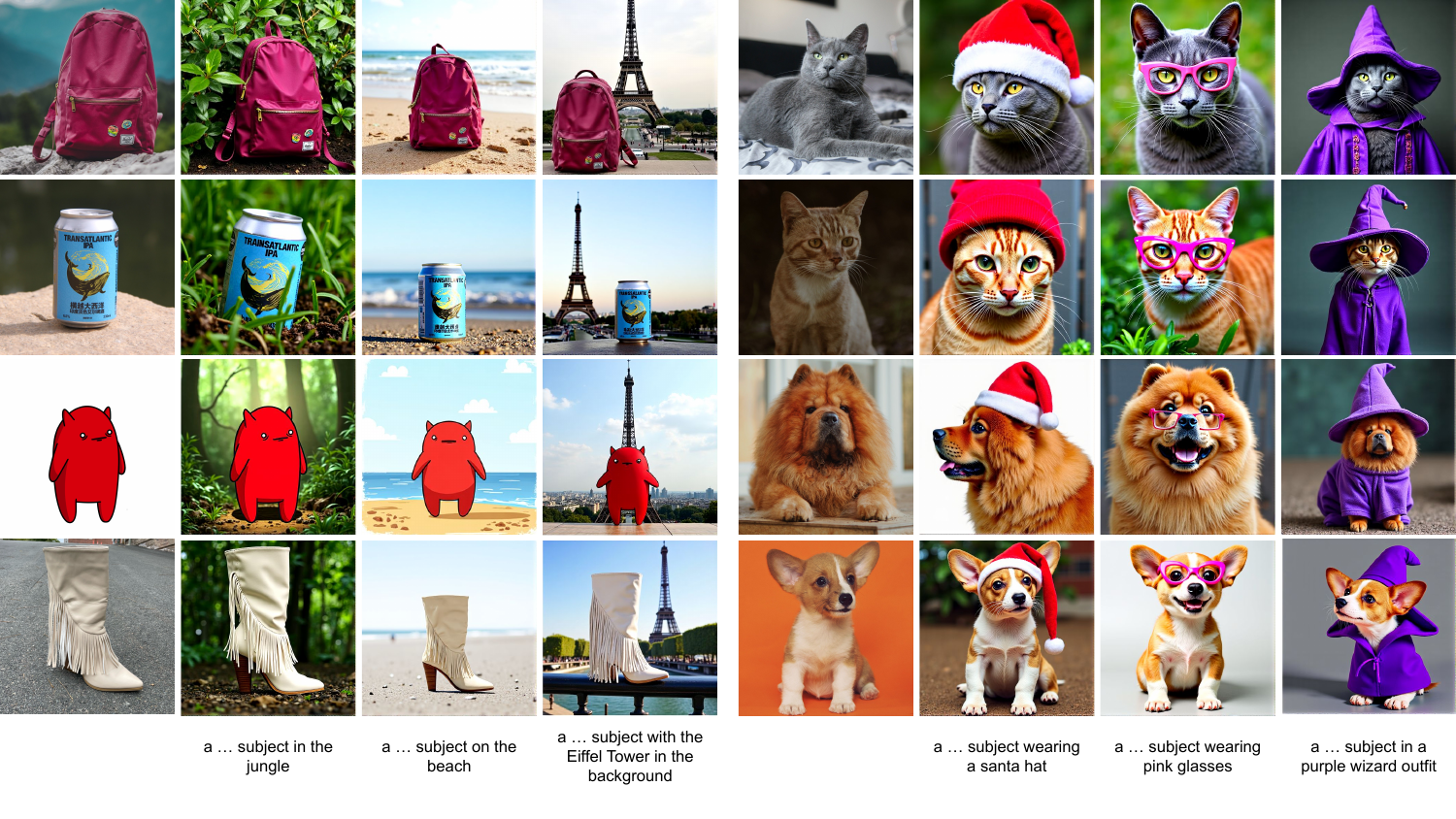}
    \caption{More qualitative results on DreamBooth Dataset.}
    \label{fig:qualitative}
\end{figure*}

We present additional qualitative results from our main experiments on the DreamBooth Dataset~\cite{ruiz2023dreambooth} in Figure~\ref{fig:qualitative}.

\subsection{More on Applications}
\begin{figure*}[th]
  \centering
  \makebox[\textwidth]{\includegraphics[width=0.8\textwidth, trim=0 125 0 0, clip]{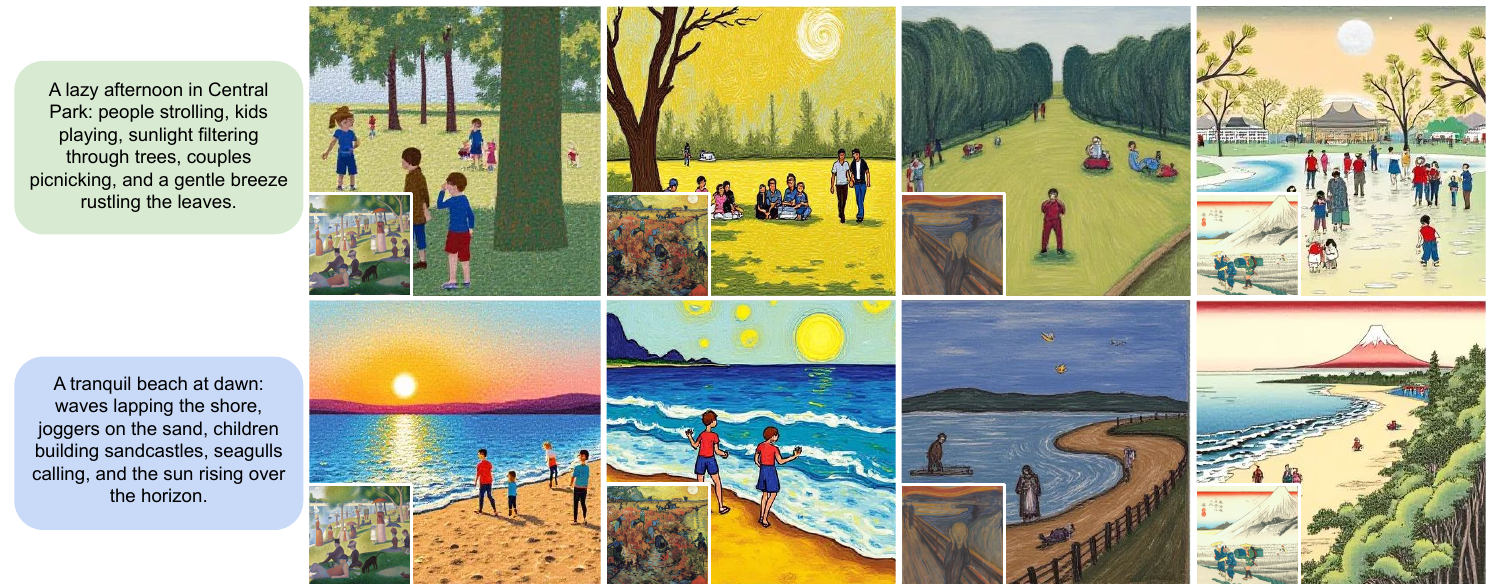}}
  \caption{Our method supports style transfer tasks. We test styles using \textit{A Sunday on La Grande Jatte} by Georges Seurat, \textit{Red Vineyards} by Vincent van Gogh, \textit{The Scream} by Edvard Munch, and \textit{Hara} by Utagawa Hiroshige.}
  \label{fig:style}
\end{figure*}

\begin{figure*}[th]
  \centering
  \makebox[\textwidth]{\includegraphics[width=0.8\textwidth, trim=0 233 0 0, clip]{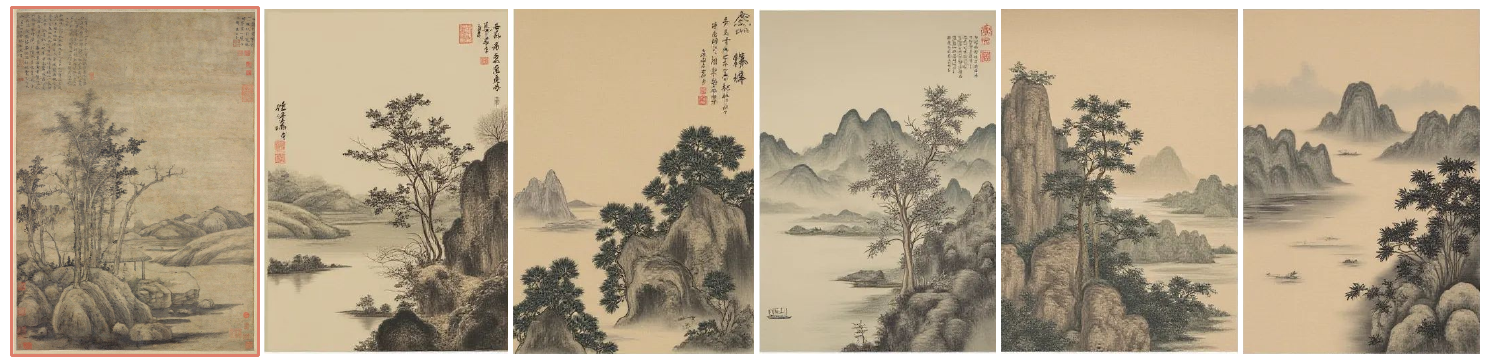}}
  \caption{Additional demonstrations of style transfer tasks are provided, showcasing the application of the Chinese ink wash painting style. We test the style using \textit{Enjoying the Wilderness in an Autumn Grove} by Ni Zan with the prompt: "Mountains, trees, and creek." The leftmost image is the reference style image. }
  \label{fig:style2}
\end{figure*}

Beyond subject-driven generation tasks, our method also supports style transfer by adjusting prompts toward a desired style, as demonstrated in Figure~\ref{fig:style}. However, we observed that the results are of higher quality when the prompt content aligns closely with the reference image. For instance, transferring Van Gogh's Sunflower style to a scene in Central Park does not yield satisfactory results. We also notice slight declines in text alignment with style transfer tasks. Chinese ink wash painting aims to capture the perceived "spirit" or "essence" of a subject rather than directly imitating it, which often does not align with semantic structures, making style transfer challenging for many reference images. However, our method performs well with Ni Zan's artwork when the text prompt includes common ink wash painting elements, such as mountains, trees, and creeks, which are frequently depicted in these artworks, as demonstrated in Figure~\ref{fig:style2}.

\end{document}